# When Does Training on synthetic data provably work?

**Leroy Odunlami** **Tobias Riedlinger** **Hanno Gottschalk**

Institute of Mathematics, Technische Universität Berlin
Berlin, Germany

## Abstract

High-quality training data are a major bottleneck in machine learning, motivating increasing use of generated data. We study when a binary classifier trained on synthetic data transfers successfully to a real distribution, under a shared conditional label distribution. First, we derive finite-sample excess-error bounds when the real covariate distribution is absolutely continuous with respect to the synthetic distribution under a controlled density ratio. Second, we look at the partially and fully singular case, where the real distribution contains mass outside the synthetic distribution. In this scenario, if the hypothesis class is sufficiently expressive, synthetic risk alone cannot uniformly control real-world error: a classifier can be arbitrarily close to optimal on synthetic data while approaching worst-case error on the uncovered real region, even when the distributions are arbitrarily close in Wasserstein distance. For a restricted function class, where the hypotheses and the log-odds of the conditional label probability satisfy a Lipschitz condition, we recover a bound in terms of the probability that a real sample lies outside the reach of the synthetic training data, restoring a transfer guarantee even in the fully singular case. We confirm our results experimentally in controlled low-dimensional settings as well as on an application-scale dataset: Using 32,200 images generated as counterparts to real ImageNet images from four different state-of-the-art image generators, we show the existence of an adversarial classifier that exhibits up to 0.5% error rate on the synthetic domain while degenerating to 99.25% error on the real domain. At the same time, across a diverse collection of ordinarily trained architectures, synthetic and real test errors remain positively correlated. These results distinguish the worst possible case from typical empirical behavior. Our results suggest that synthetic data should be budgeted at very high sample counts and models trained thereon should be rigorously validated on real data.

## 1 Introduction

Obtaining high-quality training data in sufficient quantities is one of the biggest challenges during the development of better machine learning models and can be prohibitively expensive for many use-cases Richter et al. (2016),Cordts et al. (2016). Also, labeling through human workers which is required for many applications is error-prone Northcutt et al. (2021) due to the tedious and time-consuming nature of the task. Rapid progress in generative modeling raises the question whether artificially generated (synthetic) data can replace real data in training, without compromising safety Bauer et al. (2024), Lu et al. (2025). Other potential use cases include addressing class imbalances by augmenting underrepresented categories Sampath et al. (2021), as well as enhancing privacy by substituting personal and sensitive data with synthetic alternatives. The use of synthetic data, however, introduces a *domain gap* Schwonberg et al. (2023) that must be carefully evaluated in order to determine whether supplementation with synthetic data is a viable approach for a specific task. Despite the surge in synthetic data for training, existing theory focuses on conditions under which synthetic training succeeds, while the question of what can go wrong has received little attention. In order to close this gap in the literature, we will model the situation in the framework of probability theory in order to derive excess-error bounds. We investigate a binary classification problem, where covariates are drawn from a real distribution $P_X$ and a synthetic distribution $Q_X$. Labels on the other hand will be drawn according to a shared conditional label distribution (*concept map*), yielding joint distributions $P$ and $Q$. Based on this view point, we identify 3 different scenarios that govern synthetic training: The absolutely continuous case, the fully singular case and the mixed case. A Hölder-type bound on the Radon-Nikodym derivative allows us to transfer rates of convergence from the classical case to the absolutely continuous case. In the mixed case, where the synthetic distribution covers the real one only partially, synthetic training leaves the uncovered region unconstrained: the empirical synthetic risk cannot distinguish the Bayes rule from an adversarial one. In the singular case, this construction affects all of $P_X$ and there exists a model that has optimal accuracy under $Q$ and worst possible accuracy under $P$. This negative result relies on the existence of a hypothesis that switches abruptly between the two supports. A Lipschitz condition on the model class and on the conditional label distribution rules such hypotheses out and restores a guarantee. Our contributions are the following:

1. We show that if $P_X$ is absolutely continuous with respect to $Q_X$ with density ratio in $L^s(Q_X)$, classical statistical learning rates transfer to synthetic training (Theorem 3).
2. We show that otherwise, under the Lebesgue decomposition theorem, sufficiently rich classifiers can achieve $\varepsilon$-small error under $Q$ while being $\varepsilon$-close to worst possible accuracy on the singular component of $P_X$ (Theorem 4).
3. In the fully singular case, we prove the existence of an adversarial classifier which is equal to the Bayes classifier on the support of $Q_X$ and opposite to it on the support of $P_X$ (Corollary 5). Further, we show that Lipschitz constraints on the model class and the concept map bound the excess-error by the probability that a real sample falls outside the reach of the synthetic support (Theorem 7).
4. We observe the above phenomena numerically in both a controlled low-dimensional setting and a realistic high-dimensional setting and provide empirical evidence that partial training success in the sense of Theorem 4 is typical.

The paper is structured as follows: Section 2 gives an overview of related literature. In Section 3 we derive convergence rates for classifiers trained on synthetic data where possible and identify scenarios where this is not possible. Section 4 demonstrates our findings experimentally in controlled low-dimensional settings as well as using state-of-the-art generative models.

## 2 Related Work

Synthetic training data is used across disciplines. Diffusion-generated images improve ImageNet classification (Azizi et al., 2023), models trained on ImageNet replicas perform competitively (Sariyildiz et al., 2023), brain MRI segmentation networks can be trained on synthetic scans alone (Billot et al., 2023), and small language models trained on synthetic text data match larger ones on coding benchmarks (Gunasekar et al., 2024). A smaller line of research examines the risks: recursive training on model outputs erases the tails of the original distribution (Shumailov et al., 2024).

The theoretical underpinnings of training under distributional mismatch are rooted in the covariate-shift literature. Shimodaira (2000) shows that density-ratio reweighting is asymptotically optimal under model misspecification. Such reweighting is well-defined only when the test distribution is absolutely continuous with respect to the training distribution, and breaks down otherwise. David et al. (2010) prove impossibility results showing that the covariate-shift assumption alone is insufficient for guaranteeing learning on the target domain. Ben-David et al. (2010) derive the canonical domain-adaptation bound, decomposing target error into source error and an $\mathcal{H}\Delta\mathcal{H}$-divergence term, refined by Mansour et al. (2009) into the loss-aware discrepancy distance. Chang et al. (2019) bound the real-data risk of a classifier trained on GAN-synthetic samples, decomposing it into the synthetic risk, an ideal-joint-hypothesis error, and a refined hypothesis-pair divergence $d_{h\Delta h^*}$. Xu et al. (2023) bound the synthetic-versus-real generalization gap by feature-distribution discrepancy, conditional-label approximation error, and hypothesis-class misspecification. More recently, Nguyen et al. (2025) bound the real risk of a predictor trained on real and synthetic samples combined, via output-space discrepancy, local-robustness and a concentration term. Shidani et al. (2026) derive a stability-based bound for mixtures in which the excess risk depends on the Wasserstein-2 distance, and use it to characterize an optimal synthetic-to-real ratio, under a Lipschitz hypothesis class. Exact asymptotics are available in restricted settings (El Firdoussi et al., 2025). Unlike prior works whose bounds are mostly phrased in terms of a distance between the synthetic and real distributions, our focus is to show that for unrestricted hypothesis classes vanishing Wasserstein distance is not sufficient for convergence, i.e. the error on the real domain does not converge uniformly.

# 3 Theoretical Results

## 3.1 Notation and Assumptions

**Data and distributions.** Let $P_X$ and $Q_X$ be probability distributions on $(\mathcal{X}, \mathcal{B}(\mathcal{X}))$, where $\mathcal{X} \subset \mathbb{R}^d$ is compact and $\mathcal{B}(\mathcal{X})$ denotes the Borel $\sigma$-algebra on $\mathcal{X}$. We think of $\mathcal{X}$ as the data space, and of $P_X$ and $Q_X$ as the true distribution and the generative model. We pair $\mathcal{X}$ with the labels $\mathcal{Y} = \{0, 1\}$ and assume a true conditional distribution $p(\cdot\,|x)$ on $\mathcal{Y}$, with $\eta(x) := p(1|x)$. The joint distributions $P$ and $Q$ on $\mathcal{X} \times \mathcal{Y}$ pair each covariate distribution (i.e. distribution on $\mathcal{X}$) with this shared conditional, i.e. for $C \in \mathcal{B}(\mathcal{X} \times \mathcal{Y})$, $P(C) = \int_{\mathcal{X}} \sum_{y \in \mathcal{Y}} \mathbf{1}_C(x, y)\, p(y|x)\, P_X(\mathrm{d}x)$, and the same analogously for $Q$ with $Q_X$ in place of $P_X$. The synthetic training set $V_n$ consists of the first $n$ elements of an i.i.d. sequence $(X_i, Y_i)_{i \in \mathbb{N}} \sim Q$.

**Model.** We model $p(y|x)$ by $p_\theta(y|x) = y\,(\sigma \circ f_\theta)(x) + (1 - y)[1 - (\sigma \circ f_\theta)(x)]$, with the sigmoid $\sigma(t) = \exp(t)/(1 + \exp(t))$ and its inverse `logit`. The model is given by a compact $\Theta \subset \mathbb{R}^m$ and a map $\Phi : \Theta \times \mathcal{X} \to \mathbb{R}$ with $\|\Phi\|_\infty \leq B < \infty$, continuous in the first and measurable in the second argument. We write $f_\theta := \Phi(\theta, \cdot)$, $\mathcal{F} := \{f_\theta \mid \theta \in \Theta\}$, $\gamma := \sigma(-B) \in (0, 1/2)$ and $\eta_\theta(x) := p_\theta(1|x)$.

**Classifiers and risk.** For a measurable classifier $h : \mathcal{X} \to \{0, 1\}$ the classification error (or error rate) is $\mathrm{err}_P(h) := P(\{h(X) \neq Y\})$, minimized by the Bayes-optimal classifier $y^*(x) \in \arg\max_y p(y|x)$, which need not be of the form $\hat{y}_\theta$ and is not necessarily unique. Our goal is to train $\hat{y}_\theta(x) = \mathbf{1}\{\eta_\theta(x) > \frac{1}{2}\}$ by minimizing the empirical risk $\hat{R}(\theta, V_n) = \frac{1}{n} \sum_{(x,y) \in V_n} \ell(\theta, x, y)$ over $\theta \in \Theta$, for the negative log-likelihood $\ell(\theta, x, y) = -\log p_\theta(y|x)$ with population risk $R_Q(\theta) := \mathbb{E}_Q[\ell(\theta, X, Y)]$. We write $\hat{\theta}_n \in \arg\min_{\theta \in \Theta} \hat{R}(\theta, V_n)$, $\theta^* \in \arg\min_{\theta \in \Theta} R_Q(\theta)$ and $\varepsilon_{\mathcal{F}} := \mathbb{E}_{X \sim Q_X}[\mathrm{KL}_X(p \,\|\, p_{\theta^*})]$ for the approximation floor of $\mathcal{F}$. Both are attained, since $\ell(\theta, x, y) \in \left[\log \frac{1}{1-\gamma}, \log \frac{1}{\gamma}\right]$ is continuous in $\theta$, so $R_Q$ and $\hat{R}(\cdot, V_n)$ are continuous on the compact $\Theta$.

**Divergences and complexity.** For distributions $\mu, \nu$ on $\mathcal{Y}$ with $\mu \ll \nu$ we write $\mathrm{KL}(\mu \,\|\, \nu) := \sum_y \mu(y) \log(\mu(y)/\nu(y))$ and $H(\mu, \nu) := -\sum_y \mu(y) \log \nu(y)$, with $H(\mu) := H(\mu, \mu)$, and abbreviate $\mathrm{KL}_x(\mu \,\|\, \nu) := \mathrm{KL}(\mu(\cdot\,|x) \,\|\, \nu(\cdot\,|x))$, likewise $H_x$. Since $\mathcal{Y} = \{0, 1\}$, $\mathrm{KL}_x(p \,\|\, p_\theta) = \mathrm{KL}\big(\mathrm{Bern}_{\eta(x)} \| \mathrm{Bern}_{\eta_\theta(x)}\big)$ for the Bernoulli distribution $\mathrm{Bern}_a$ with parameter $a$. For $\mu, \nu$ on $\mathcal{X}$ we use the total variation distance $\mathrm{TV}(\mu, \nu) = \sup_{A \in \mathcal{B}(\mathcal{X})} |\mu(A) - \nu(A)|$ and the Wasserstein-1 distance $W_1(\mu, \nu) := \inf_{\pi \in \Gamma(\mu, \nu)} \mathbb{E}_{(x, x') \sim \pi}[d(x, x')]$, where $\Gamma(\mu, \nu)$ is the set of couplings of $\mu$ and $\nu$ and $d$ the Euclidean metric. Finally, Dudley's entropy integral for the loss class $\mathcal{L} = \{-\log p_\theta(\cdot \mid \cdot) \mid \theta \in \Theta\}$ is $\mathcal{J}(\mathcal{L}) = \int_0^{\log((1-\gamma)/\gamma)} \sqrt{\log N(t, \mathcal{L}, \|\cdot\|_\infty)}\,\mathrm{d}t$, where $N$ denotes the covering number. We assume $\mathcal{J}(\mathcal{L}) < \infty$ throughout, Remark 2 verifies this for deep ReLU networks.

**Assumptions.** Each of our results relies on a subset of the following assumptions:

- $(L_p)$ $p(y|x)$ is Lipschitz in $x$ with constant $L_p$.
- $(L^*)$ $f^* := \mathtt{logit} \circ \eta$ is Lipschitz with constant $L^*$.
- $(\kappa)$ $\exists \kappa > 0$ with $p(y|x) > \kappa$ for all $x, y$. Set $F := \log((1 - \kappa)/\kappa)$.
- $(L^\infty)$ $P_X \ll Q_X$ and $\|\mathrm{d}P_X/\mathrm{d}Q_X\|_{L^\infty(\mathcal{X}, Q_X)}$ is finite.
- $(L^s)$ $P_X \ll Q_X$ and $\|\mathrm{d}P_X/\mathrm{d}Q_X\|_{L^s(\mathcal{X}, Q_X)}$ is finite, $s \in (1, \infty)$.
- $(L_f)$ Every $f \in \mathcal{F}$ is Lipschitz with $L_f := \sup_{f \in \mathcal{F}} \mathrm{Lip}(f) < \infty$.

## 3.2 Classical Setting

The following analysis is organized into 3 cases according to the relationship between the real and synthetic distribution: absolute continuity ($P \ll Q$), partial absolute continuity ($P \not\ll Q$) and the singular case ($P \perp Q$), cf. Figure 1. As a baseline, we will first look at the standard supervised learning case, where training and test data are sourced from the same distribution. Rates of this type are known in the literature under varying sets of assumptions, cf. Bartlett et al. (2006); Zhang et al. (2024). Here, we prove a bound on the excess classification error that holds for every model class in the sense specified above, with no additional architectural assumptions. Throughout, $\lesssim$ denotes inequality up to a multiplicative constant.

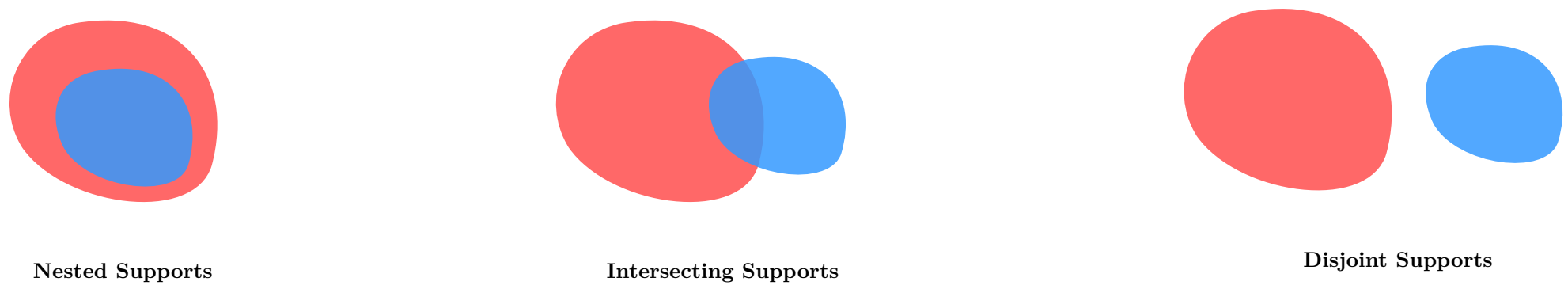

Figure 1: Example for the 3 possible cases (left to right): Absolute continuous case, mixed case and singular case

**Theorem 1** (Convergence rate under matched distribution). *For any $\delta \in (0,1)$ it holds that $\mathrm{err}_Q\big(\hat{y}_{\hat{\theta}_n}\big) - \mathrm{err}_Q(y^*) \lesssim \sqrt{\varepsilon_{\mathcal{F}} + r_n(\delta)}$ with $r_n(\delta) = \Big[\mathcal{J}(\mathcal{L}) + \log(1/\gamma)\Big(1 + \sqrt{\log(2/\delta)}\Big)\Big]/\sqrt{n}$ with probability at least $1 - \delta$. (Proof in Appendix A.1)*

**Remark 2.** The quantities in Theorem 1 can be made more concrete for specific architectures. For instance under ($\kappa$) for truncated deep ReLU networks of depth $L$ and weight bound $W$, one obtains $\varepsilon_{\mathcal{F}} \leq \mathrm{dist}^2_\infty(f^*, \mathcal{F})/(16\kappa(1-\kappa))$ with $f^* := \mathtt{logit} \circ \eta$ and $\mathcal{J}(\mathcal{L}) \lesssim L\sqrt{\log W}$. (Statement and proof in Appendix A.2)

We now turn to the setting of two different distributions $P_X$ and $Q_X$. In general, the generator will cover a part of the real distribution and miss the rest. Formally, Lebesgue's decomposition splits $P_X$ into an absolutely continuous and a singular part with regard to $Q_X$, i.e. $P_X = P_X^a + P_X^s$ with $P_X^a \ll Q_X$ and $P_X^s \perp Q_X$. Moreover, there is a partition $\mathcal{X} = M_a \sqcup M_s$ with $Q_X(M_s) = 0$ such that $P_X^a = P_X(\,\cdot \cap M_a)$ and $P_X^s = P_X(\,\cdot \cap M_s)$. Let $P^a := P(\,\cdot \cap (M_a \times \mathcal{Y}))$ and $P^s := P(\,\cdot \cap (M_s \times \mathcal{Y}))$ be the corresponding parts of the joint distribution, so that $P = P^a + P^s$. The target error splits[1] accordingly into $\mathrm{err}_P = \mathrm{err}_{P^a} + \mathrm{err}_{P^s}$. The following sections investigate the different scenarios: $P = P^a$, i.e. the generator captures the whole real distribution, $P = P^s$, i.e. the generator misses the entire distribution and the general case $P = P^a + P^s$, where $P^a, P^s \neq 0$. We will see that in the absolutely continuous case (Section 3.3), the derived results carry over. In the non-absolutely continuous case (Sections 3.4 and 3.5), the results do not carry over absent further assumptions.

### 3.3 Case 1: $P$ is absolutely continuous with regard to $Q$

**Theorem 3** (Transfer of rates). *For each $\delta \in (0,1)$ with probability at least $1-\delta$:*

*(i) Under ($L^\infty$), it holds that $\mathrm{err}_P\big(\hat{y}_{\hat{\theta}_n}\big) - \mathrm{err}_P(y^*) \lesssim \sqrt{\varepsilon_{\mathcal{F}} + r_n(\delta)}$*

*(ii) Under ($L^s$), it holds that $\mathrm{err}_P\big(\hat{y}_{\hat{\theta}_n}\big) - \mathrm{err}_P(y^*) \lesssim (\varepsilon_{\mathcal{F}} + r_n(\delta))^{\frac{1}{2t}}$,*

*where $\frac{1}{s} + \frac{1}{t} = 1$ and $r_n(\delta) := \Big[\mathcal{J}(\mathcal{L}) + \log(1/\gamma)\Big(1 + \sqrt{\log(2/\delta)}\Big)\Big]/\sqrt{n}$. (Proof in Section A.3)*

This shows that under above regularity assumptions the rates do carry cover. Regarding the plausibility of these assumptions, absolute continuity fails as soon as $Q_X$ assigns zero probability to an event where $P_X$ assigns positive probability. This can easily happen, e.g. if the supports of both distributions have different intrinsic dimension. For generative models such as variational autoencoders (Kingma & Welling, 2022) and generative adversarial networks (Goodfellow et al., 2014), which map a low-dimensional latent space into the data space, the synthetic support will most likely have a smaller dimension compared to the real one. For models such as normalizing flows (Rezende & Mohamed, 2015), neural ODEs (Chen et al., 2018), diffusion models (Ho et al., 2020), and flow matching (Lipman et al., 2023), whose induced distributions may have positive density over the entire space, the situation is less clear. In either case, the absolutely continuous case can hardly be verified in realistic settings. Next, we will look at the general case where $P$ is partially absolutely continuous, partially singular with regard to $Q$.

[1]Since $P^a$ and $P^s$ are not probability measures, $\mathrm{err}_{P^a}$ and $\mathrm{err}_{P^s}$ are not error rates in $[0,1]$ but misclassified probability mass, bounded by $P_X(M_a)$ and $P_X(M_s)$, respectively.

### 3.4 Case 2: $P$ splits into an absolutely continuous and a singular part

**Theorem 4** (Higher expressivity of $\mathcal{F}$ leads to higher worst-case error)**.** *Let $\varepsilon > 0$ and set $\tilde{y} := y^* \mathbf{1}_{M_a} + \hat{y}_{\text{adv}} \mathbf{1}_{M_s}$ with $\hat{y}_{\text{adv}} := \mathbf{1}\{\eta < \frac{1}{2}\}$. Suppose there exists $\theta \in \Theta$ with $P_X(\hat{y}_\theta \neq \tilde{y}) \leq \varepsilon$ and $Q_X(\hat{y}_\theta \neq \tilde{y}) \leq \varepsilon$. Then $\text{err}_Q(\hat{y}_\theta) - \text{err}_Q(y^*) \leq \varepsilon$, but $\text{err}_{P^s}(\hat{y}_\theta) \geq \max_h \text{err}_{P^s}(h) - \varepsilon$ the maximum being over all measurable classifiers $h$. (Proof in Section A.4)*

Theorem 4 shows the more expressive our function class is with regard to our learning target the higher the worst-case error. The assumptions $P_X(\hat{y}_\theta \neq \tilde{y}) \leq \varepsilon$ and $Q_X(\hat{y}_\theta \neq \tilde{y}) \leq \varepsilon$ encode the expressivity of the model class. They are strictly weaker than a universal approximation property since they require $\varepsilon$-approximation of the given target $\tilde{y}$ only rather than a whole function space. The theorem does not assert that empirical risk minimization will select such a parameter. Minimizing the synthetic risk however, cannot rule out this scenario, not even with an infinite amount of synthetic training data. Since the adversarial and the Bayes-optimal behavior differ only on a $Q_X$-null set, no criterion computable from synthetic data distinguishes them. Which minimizer a concrete training pipeline returns is determined by inductive biases in the model class such as architecture, initialization as well as optimization algorithm that are independent of the probability distributions of the data.

### 3.5 Case 3: $P$ is completely singular to $Q$

We now consider the case of completely singular distributions. Under $P_X \perp Q_X$ the synthetic distribution leaves no part of the target covered, so the adversarial construction of Theorem 4 affects all of $P$ rather than a part of it. This is due to the freedom of the hypothesis class to behave arbitrarily where no synthetic data lies. No criterion computable from synthetic data can exclude such a classifier, so without further conditions on the hypothesis space or the optimization algorithm, no uniform control over the target error is possible in this regime.

**Corollary 5** (Adversarial classifier)**.** *Assume $P_X \perp Q_X$ in addition to the hypotheses of Theorem 4. Then there exists $\theta \in \Theta$ with $\text{err}_Q(\hat{y}_\theta) - \text{err}_Q(y^*) \leq \varepsilon$, but $\text{err}_P(\hat{y}_\theta) \geq 1 - \text{err}_P(y^*) - \varepsilon$.*

*Proof.* We have $P^a = 0$ and $P^s = P$. Hence, $\text{err}_P = \text{err}_{P^s}$, and Theorem 4 yields $\theta \in \Theta$ with the stated bound under $Q$ and $\text{err}_P(\hat{y}_\theta) \geq \max_h \text{err}_P(h) - \varepsilon$. As shown in the proof of Theorem 4, $\max_h \text{err}_P(h) = \mathbb{E}_P\left[\max_y p(y|X)\right]$. Since $\mathcal{Y} = \{0, 1\}$, we have $\max_y p(y|x) + \min_y p(y|x) = 1$ for every $x$, and $\text{err}_P(y^*) = \mathbb{E}_P\left[\min_y p(y|X)\right]$, so $\max_h \text{err}_P(h) = 1 - \text{err}_P(y^*)$. ☐

For illustrative purposes, consider the case where the Bayes error is zero, i.e. perfect classification is possible. Then the equation instantiates to $\text{err}_P(\hat{y}_\theta) \geq 1 - \varepsilon$, i.e. the error rate comes arbitrarily close to 1, which is the worst case possible. As already in the previous case, this result rules out uniform consistency under synthetic training. Implications of this will be discussed in section Section 5. Existing bounds are often phrased in terms of Wasserstein distance between the real and the synthetic distribution. A divergence however, e.g. Wasserstein distance may be arbitrarily small even for mutually singular distributions, in which case our negative results apply. Whether the supports are truly disjoint is hard to verify. One can however translate the abstract singularity condition into a more intuitive proxy criterion which can be tested in practice, see Section 4.3.

**Remark 6.** $P_X$ and $Q_X$ are mutually singular if and only if there exists a measurable $D : \mathcal{X} \to \{0, 1\}$ that separates the two distributions perfectly at the population level, i.e. $P_X(D = 1) = 1$ and $Q_X(D = 1) = 0$.

*Proof.* If $P_X$ and $Q_X$ are mutually singular, there exists $A \in \mathcal{B}(\mathcal{X})$ with $P_X(A) = 1$ and $Q_X(A) = 0$, and $D = \mathbf{1}_A$ separates perfectly. Conversely, given such a $D$, the set $A := D^{-1}(\{1\})$ satisfies $P_X(A) = 1$ and $Q_X(A) = 0$, which is the definition of mutual singularity. ☐

Although this is merely a proxy criterion and not a mathematically rigorous check of singularity, it gives us an idea to which extent the distributions appear to be singular from the viewpoint of the neural network. Constraining the hypothesis class through a Lipschitz condition restores upper bounds on the classification error. The intuition behind this construction is simple: functions that cannot change arbitrarily fast will assume similar values on nearby regions.

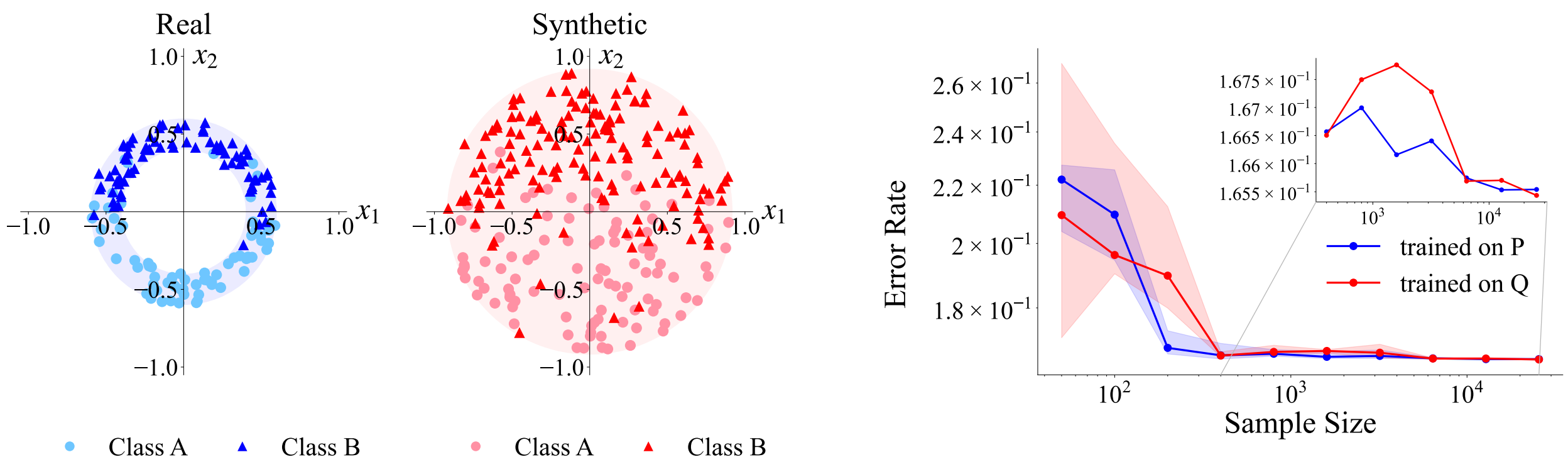

Figure 2: Real (left) and synthetic (middle) distribution and convergence of error rates (right). The left and middle plot show a true (blue) distribution with ring-shaped support and a synthetic (red) distribution with disk-shaped support covering the real one. The right plot shows the achieved classification error on a (real/blue) test data set after a full training run on a training set of size n.

**Theorem 7** (Lipschitz bound). *Assume* ($\kappa$), ($L_f$) *and* ($L^*$), *and let* $y^*(x) = \mathbf{1}\{f^*(x) > 0\}$. *Define* $r_\theta(x) := \min\{|f_\theta(x)| \,/L_f, |f^*(x)| \,/L^*\}$, $\quad S := \{x \in \operatorname{supp} Q_X \mid \hat{y}_\theta(x) = y^*(x)\}$ $\quad$ *and* $\quad U_\theta := \bigcup_{x \in S} B(x, r_\theta(x))$. *Then* $\operatorname{err}_P(\hat{y}_\theta) - \operatorname{err}_P(y^*) \leq P_X(X \notin U_\theta)$. *In particular, if* $\operatorname{supp} P_X \subseteq U_\theta$, *then* $\operatorname{err}_P(\hat{y}_\theta) = \operatorname{err}_P(y^*)$. *(Proof in Appendix A.6)*

**Remark 8.** For a bound that is in spirit more similar to Theorem 1 and Theorem 3, cf. Appendix B.3.

## 4 Experiments

We will now check our most important theoretical results through a series of experiments. First, we use simple low-dimensional distributions allowing us to control all relevant quantities. This way, we ensure that all assumptions of the theorems are met and that the underlying effects can be observed. We then scale the experiments up to a realistic synthetic dataset using multiple different state-of-the-art generative image models. Our results show that in practice the disjoint support scenario is relevant and synthetically trained models need to be rigorously validated against real data, especially for safety-relevant applications.

### 4.1 Absolutely continuous case

Theorem 3 established a transfer of convergence rates up to a multiplicative factor in the case where $P_X \ll Q_X$. To validate this in a controlled setting, we drew samples from uniform distributions on the disk and on an annulus contained therein, cf. Figure 2. Subsequently, we trained Multilayer Perceptrons (MLPs) on real and synthetic data and evaluated on a held-out real dataset. The lines in Figure 2 (right) show the median over 100 seeds and the band the interquartile range. For more details confer Appendix B.1.

The plot shows that both models converge at a comparable rate. The real median appears to converge slightly faster, which is to be expected, but the effect is not strongly visible against the seed noise. What's more distinct is that the synthetic model's variance is much higher at lower sample size. This is unsurprising given that by coincidence the coverage of the annulus by red samples can be very sparse. Hence, it makes sense that the performance of a model trained on any given low sample count synthetic training set would strongly depend on chance.

Both effects, larger variance at small training set sizes as well slower convergence speed suggest that synthetic data should be budgeted at a much higher sample count than its real counterpart. This makes intuitive sense given that a higher synthetic sample count increases the probability that some of the samples will resemble real data closely enough. Inflating the size of a synthetic training set in order to compensate for these effects should be unproblematic as the appeal of synthetic data is exactly its low marginal cost and better availability. It is however an effect worth accounting for.

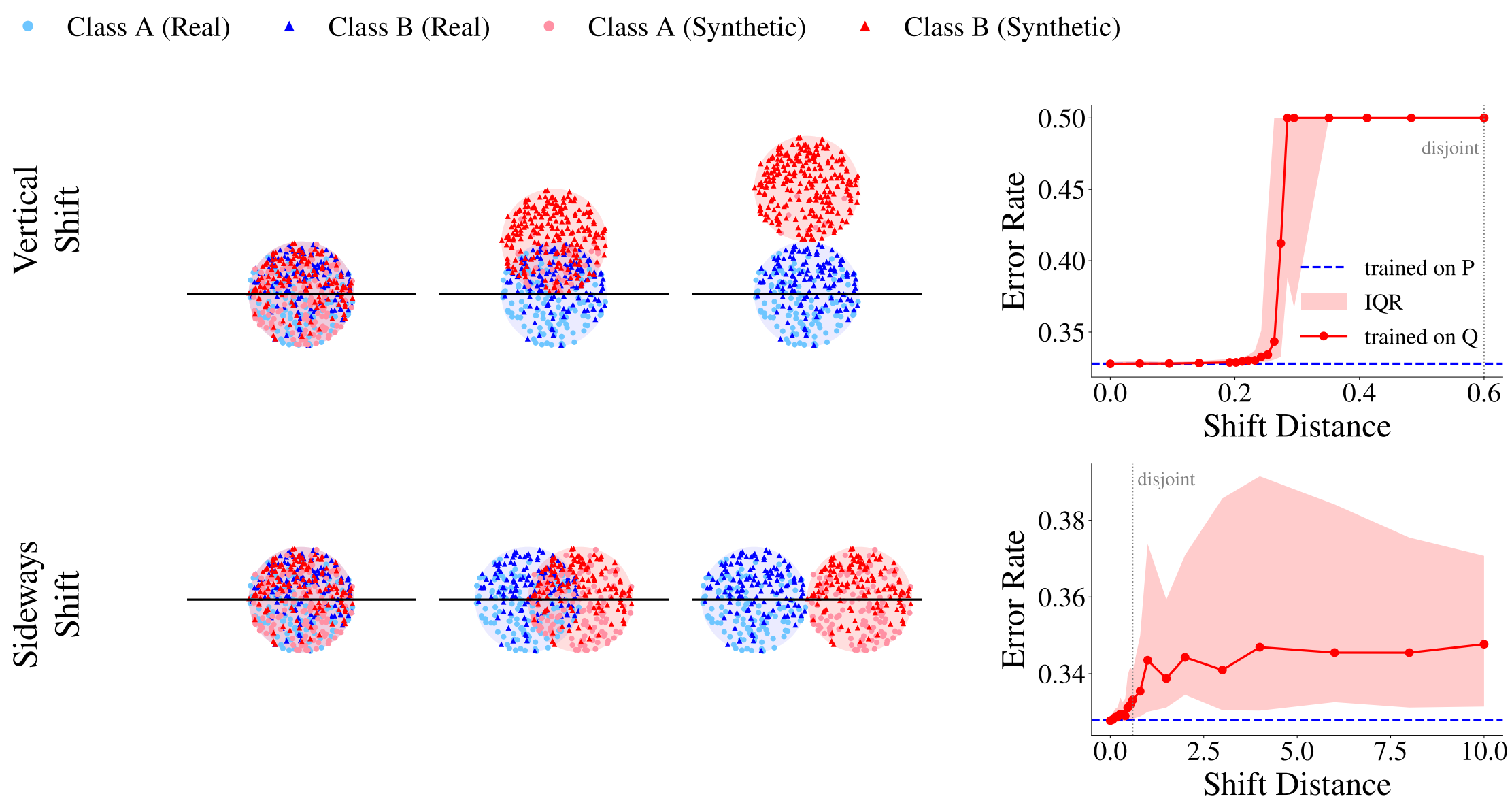

Figure 3: Datasets (left) and error rates vs. shift distance (right).

### 4.2 Mixed absolutely continuous and singular case

Theorem 4 and Corollary 5 show that adversarial networks exist, not that training finds them. To measure both the typical and the worst case, we start from identical distributions and gradually move the synthetic support away from the real one, so that the test error can be monitored as a function of displacement alone, cf. Figure 3. We use two paths: vertical displacement, across the decision boundary, and horizontal displacement, along it.

Along the vertical path the error leaves the Bayes rate steeply and approaches random guessing almost as a step function: as the synthetic disk moves up, class A disappears from the training set and the model has no chance of recovering the concept map. Near shift 0.225 the spread is remarkable, with the best runs still at the Bayes error and the worst already at chance. The horizontal path keeps the classes balanced, and accordingly the synthetic median stays well below chance up to the 25th percentile even at full disjointness. The variance across runs nonetheless remains large, even in this benign setting with fixed label map, balanced classes and a linear decision boundary.

The takeaway is that success of synthetic training is possible but depends on the arrangement of the supports relative to the concept map, which in practice is unknown, and that convergence on synthetic data is uninformative thereof. Finally, the worst case is easy to exhibit: in the disjoint horizontal configuration we add real samples with inverted labels to the synthetic training set. The resulting classifier lands within 0.17 percentage points of the Bayes error on synthetic and 0.16 points of the adversarial error on real data. We repeat this construction on realistic data in the next section. For our experimental validation of Theorem 7, see Appendix B.4.

### 4.3 Realistic synthetic data

In order to move from controlled model distributions to real-world distributions, we created a dataset using 4 modern generative models: FLUX.2 [dev] (Black Forest Labs), Stable Diffusion 3.5 Large (Stability AI), Qwen-Image (Alibaba), HunyuanImage 3.0 (Tencent). The purpose of this dataset is to isolate the domain gap by equating the image contents semantically. We therefore recreate an existing set of real images rather than generating new scenes from scratch, We selected eight ImageNet categories, four types of big cat and four types of bear, with 1,300 images each, yielding 10,400 real images in total, cf. Figure 4. Each image was captioned by the vision-language model *Qwen3-VL-8B-Instruct*, and the resulting captions were used as prompts for the generators. Every generator reproduced the full set of 10,400 images, using about 1,075 GPU hours (Nvidia A100 80GB) with the exception of Hunyuan Image 3.0, for which we restricted the generation to a subset of 1000 images due to its high computational cost. Per-generator comparisons involving Hunyuan Image 3.0 should be read with care, since its reduced sample count might be a confounding factor.

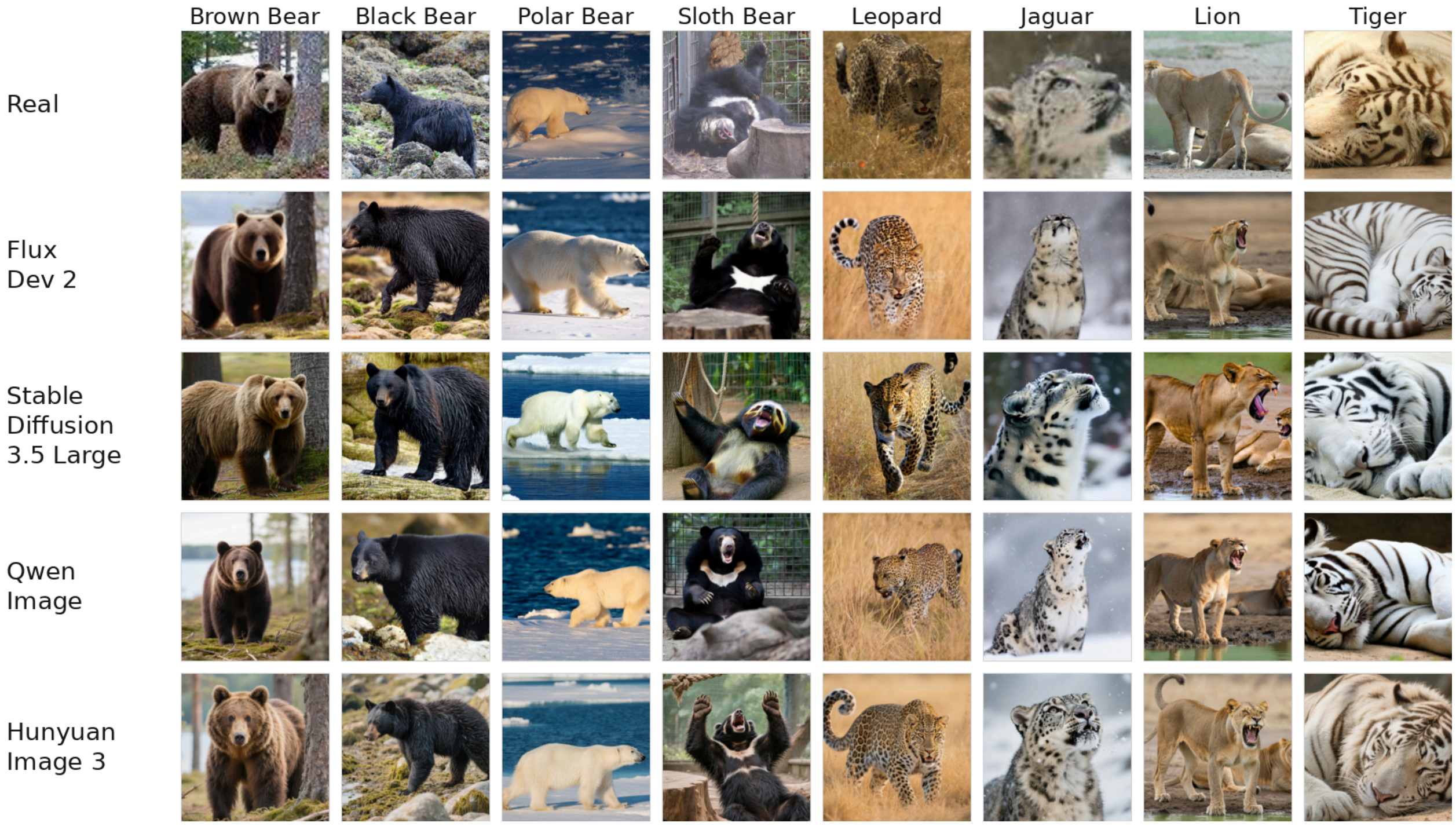

Figure 4: We used 4 state-of-the-art generative models to recreate 8 complete ImageNet classes. Descriptions of the real images were created using a vision-language model and used as prompts for the generative models. In total, 32,200 images were created, see Section 4.3.

We ran a total of 3 experiments on this data, aimed at Remark 6 and Corollary 5 as well as to relate the typical synthetic error to its real counterpart. Remark 6 noted the equivalence of the existence of a perfect real vs synthetic classifier and mutual singularity of the underlying distributions. In order to assess to what extent this is relevant in practice, we trained a model on synthetic vs real classification. For every generator, we computed the classification accuracy within every ImageNet class and then took the average across all classes. As architecture, we choose V-JEPA ViT-H/16 Bardes et al. (2024), because its training set does not include ImageNet so data leakage is ruled out. The results are depicted in Figure 5 (left). As evident from the histogram, the classifier is able to separate real and synthetic domain at a very high confidence level. Although the classification accuracy is not 100% it still suggests that synthetic data does not truly capture the real distributions support. This is surprising, given that many of the generative images are photorealistic. In contrast to the human eye, it is very easy for a deep neural network to tell real and synthetic images apart, even with state-of-the-art generative models.

Repeating the experiment around the adversarial classifier from Section 4.2 we trained an adversarial classifier on correctly labeled synthetic and inversely labeled real data. The results are depicted in Figure 5 (right). Even though the true theoretical worst case of 100% real error and 0% synthetic error are not reached, this still confirms the existence of strongly adversarial networks. The poor performance on the real domain would be invisible to any model selection process depending only on synthetic data. Note that these models would almost certainly never arise during a normal training run. This merely demonstrates existence. If $P_X \ll Q_X$, such a network should not persist as the synthetic sample size grows because the correct and the inverted labels would conflict, so no model could achieve low synthetic and high real error simultaneously. This strengthens the evidence that the supports of real and synthetic data might be mostly separated in practice, even with modern generative models. This stresses the importance of rigorous validation with real data.

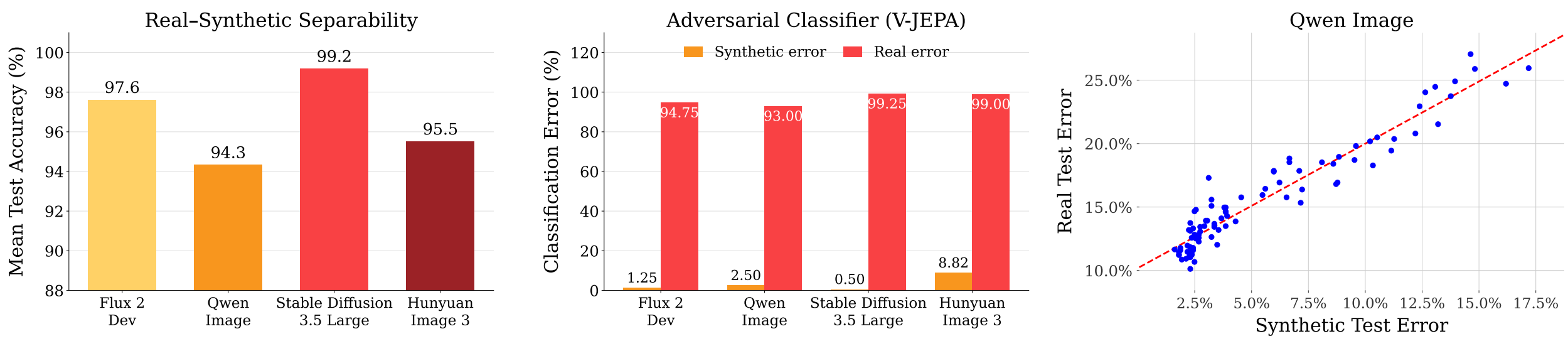

Figure 5: The left plot shows the real vs synthetic classification accuracy of a V-JEPA based classifier applied to our dataset. The middle plot shows existence of adversarial classifiers. The right plot shows correlation of real and synthetic error.

To conclude on an optimistic note, in practice the real and synthetic error appear to be somewhat predictive of one another. The worst case which is the adversarial example is possible but unlikely and this seems to hold irrespective of the specific architecture. To check this we tested a zoo of 90 different architectures from 14 families. The tested models include variants of vanilla CNNs Lecun et al. (1998), ResNet He et al. (2016) and Inception Szegedy et al. (2015), where hyperparameters like kernel size or number of layers were varied, cf. Appendix B.2 All models were trained on synthetic data alone and evaluated on held-out test sets like before. Within each generator the two errors are strongly linearly related ($r^2$ up to 0.92 (Qwen Image), $n = 90$ each). That means that in practice for modern generative models the synthetic error at least directionally indicates the magnitude of the real test error. Given that our prior experiments all provided evidence for the fact that synthetic and real data probably have mostly separated supports this fact cannot be explained through the data alone. This is only explainable due to inductive biases of deep neural networks.

## 5 Discussion

In this paper, we have proposed a theoretical framework for supervised learning under a domain gap between a synthetic and real covariate distribution. Our analysis has revealed three qualitative regimes distinguished by the relationship between the measures: absolute continuity, partial and full singularity. While our positive result in the absolutely continuous setting shows that learning rates directly transfer from synthetic to real data, the underlying assumptions cannot be rigorously verified in practice. At the same time, our experiments provide evidence that real and synthetic images can be separated with almost perfect accuracy, which suggests that the (partially) singular regime is more likely to hold in practice for the four tested generators. In this scenario, the domain gap persists even in the large-sample limit. Despite the lack of provable guarantees in this case, we typically observe a linear correlation between real and synthetic test error. Our experiments further inform how data should be allocated. Because synthetic training exhibits higher variance and slower convergence, it is advisable to use substantially larger sample sizes than real data. Real data, meanwhile, plays an irreplaceable role in validation and model selection, since synthetic test error is an unreliable proxy for real performance. Why the typical case is substantially better behaved than the worst case remains an open question. It connects to the more general question of why neural networks generalize far better than worst-case theory predicts, and deserves further investigation.

### Acknowledgement

The research leading to these results is funded by the German Federal Ministry for Economic Affairs and Energy within the project "NXT GEN AI METHODS – Generative Methoden für Perzeption, Prädiktion und Planung". The authors would like to thank the consortium for the successful cooperation.

## References

Shekoofeh Azizi, Simon Kornblith, Chitwan Saharia, Mohammad Norouzi, and David J. Fleet. Synthetic data from diffusion models improves imagenet classification. *Transactions on Machine Learning Research*, 2023. ISSN 2835-8856. URL https://openreview.net/forum?id=DlRsoxjyPm.

Adrien Bardes, Quentin Garrido, Jean Ponce, Xinlei Chen, Michael Rabbat, Yann LeCun, Mahmoud Assran, and Nicolas Ballas. Revisiting feature prediction for learning visual representations from video, 2024. URL https://arxiv.org/abs/2404.08471.

Peter L. Bartlett, Michael I. Jordan, and Jon D. McAuliffe. Convexity, classification, and risk bounds. *Journal of the American Statistical Association*, 101(473):138–156, 2006.

André Bauer, Simon Trapp, Michael Stenger, Robert Leppich, Samuel Kounev, Mark Leznik, Kyle Chard, and Ian Foster. Comprehensive exploration of synthetic data generation: A survey, 2024. URL https://arxiv.org/abs/2401.02524.

S. Ben-David, J. Blitzer, and K. et al. Crammer. A theory of learning from different domains. *Machine Learning*, 79:151–175, 2010. doi: 10.1007/s10994-009-5152-4. URL https://doi.org/10.1007/s10994-009-5152-4.

Benjamin Billot, Douglas N. Greve, Oula Puonti, Axel Thielscher, Koen Van Leemput, Bruce Fischl, Adrian V. Dalca, and Juan Eugenio Iglesias. Synthseg: Segmentation of brain mri scans of any contrast and resolution without retraining. *Medical Image Analysis*, 86:102789, 2023. ISSN 1361-8415. doi: https://doi.org/10.1016/j.media.2023.102789. URL https://www.sciencedirect.com/science/article/pii/S1361841523000506.

Fu-Chieh Chang, Hao-Jen Wang, Chun-Nan Chou, and Edward Y. Chang. G2R bound: A generalization bound for supervised learning from GAN-synthetic data. *arXiv preprint arXiv:1905.12313*, 2019. URL https://arxiv.org/abs/1905.12313.

Ricky T. Q. Chen, Yulia Rubanova, Jesse Bettencourt, and David K Duvenaud. Neural ordinary differential equations. In S. Bengio, H. Wallach, H. Larochelle, K. Grauman, N. Cesa-Bianchi, and R. Garnett (eds.), *Advances in Neural Information Processing Systems*, volume 31. Curran Associates, Inc., 2018. URL https://proceedings.neurips.cc/paper_files/paper/2018/file/69386f6bb1dfed68692a24c8686939b9-Paper.pdf.

Marius Cordts, Mohamed Omran, Sebastian Ramos, Timo Rehfeld, Markus Enzweiler, Rodrigo Benenson, Uwe Franke, Stefan Roth, and Bernt Schiele. The cityscapes dataset for semantic urban scene understanding. In *2016 IEEE Conference on Computer Vision and Pattern Recognition (CVPR)*, pp. 3213–3223, 2016. doi: 10.1109/CVPR.2016.350.

Shai Ben David, Tyler Lu, Teresa Luu, and David Pal. Impossibility theorems for domain adaptation. In Yee Whye Teh and Mike Titterington (eds.), *Proceedings of the Thirteenth International Conference on Artificial Intelligence and Statistics*, volume 9 of *Proceedings of Machine Learning Research*, pp. 129–136, Chia Laguna Resort, Sardinia, Italy, 13-15 May 2010. PMLR. URL https://proceedings.mlr.press/v9/david10a.html.

Luc Devroye, László Györfi, and Gábor Lugosi. *A Probabilistic Theory of Pattern Recognition*. Stochastic Modelling and Applied Probability. Springer New Yorks, NY, 1996. ISBN 978-0-387-94618-4. doi: 10.1007/978-1-4612-0711-5.

Alexey Dosovitskiy, Lucas Beyer, Alexander Kolesnikov, Dirk Weissenborn, Xiaohua Zhai, Thomas Unterthiner, Mostafa Dehghani, Matthias Minderer, Georg Heigold, Sylvain Gelly, Jakob Uszkoreit, and Neil Houlsby. An image is worth 16x16 words: Transformers for image recognition at scale, 2021. URL https://arxiv.org/abs/2010.11929.

Aymane El Firdoussi, Mohamed El Amine Seddik, Soufiane Hayou, Reda Alami, Ahmed Alzubaidi, and Hakim Hacid. Maximizing the potential of synthetic data: Insights from random matrix theory. In Y. Yue, A. Garg, N. Peng, F. Sha, and R. Yu (eds.), *International Conference on Learning Representations*, volume 2025, pp. 46780–46815, 2025. URL https://proceedings.iclr.cc/paper_files/paper/2025/file/7416573f05b50beac6d0aef3abc805c0-Paper-Conference.pdf.

Ian J. Goodfellow, Jean Pouget-Abadie, Mehdi Mirza, Bing Xu, David Warde-Farley, Sherjil Ozair, Aaron Courville, and Yoshua Bengio. Generative adversarial nets. In Z. Ghahramani, M. Welling, C. Cortes, N. Lawrence, and K. Weinberger (eds.), *Advances in Neural Information Processing Systems*, volume 27. Curran Associates, Inc., 2014. URL https://proceedings.neurips.cc/paper_files/paper/2014/file/f033ed80deb0234979a61f95710dbe25-Paper.pdf.

Suriya Gunasekar, Yi Zhang, Jyoti Aneja, Caio Cesar Teodoro Mendes, Allie Del Giorno, Sivakanth Gopi, Mojan Javaheripi, Piero Conti Kauffmann, Gustavo Henrique de Rosa, Olli Saarikivi, Adil Salim, Shital Shah, Harkirat Behl, Xin Wang, Sebastien Bubeck, Ronen Eldan, Adam Tauman Kalai, Yin Tat Lee, and Yuanzhi Li. Textbooks are all you need, 2024. URL https://openreview.net/forum?id=Fq8tKtjACC.

Kaiming He, Xiangyu Zhang, Shaoqing Ren, and Jian Sun. Deep residual learning for image recognition. In *2016 IEEE Conference on Computer Vision and Pattern Recognition (CVPR)*, pp. 770–778, 2016. doi: 10.1109/CVPR.2016.90.

Jonathan Ho, Ajay Jain, and Pieter Abbeel. Denoising diffusion probabilistic models. In H. Larochelle, M. Ranzato, R. Hadsell, M.F. Balcan, and H. Lin (eds.), *Advances in Neural Information Processing Systems*, volume 33, pp. 6840–6851. Curran Associates, Inc., 2020. URL https://proceedings.neurips.cc/paper_files/paper/2020/file/4c5bcfec8584af0d967f1ab10179ca4b-Paper.pdf.

Jie Hu, Li Shen, and Gang Sun. Squeeze-and-excitation networks. In *2018 IEEE/CVF Conference on Computer Vision and Pattern Recognition*, pp. 7132–7141, 2018. doi: 10.1109/CVPR.2018.00745.

Gao Huang, Zhuang Liu, Laurens Van Der Maaten, and Kilian Q. Weinberger. Densely connected convolutional networks. In *2017 IEEE Conference on Computer Vision and Pattern Recognition (CVPR)*, pp. 2261–2269, 2017. doi: 10.1109/CVPR.2017.243.

Diederik P Kingma and Max Welling. Auto-encoding variational bayes, 2022. URL https://arxiv.org/abs/1312.6114.

Y. Lecun, L. Bottou, Y. Bengio, and P. Haffner. Gradient-based learning applied to document recognition. *Proceedings of the IEEE*, 86(11):2278–2324, 1998. doi: 10.1109/5.726791.

Duo Li, Jie Hu, Changhu Wang, Xiangtai Li, Qi She, Lei Zhu, Tong Zhang, and Qifeng Chen. Involution: Inverting the inherence of convolution for visual recognition. In *2021 IEEE/CVF Conference on Computer Vision and Pattern Recognition (CVPR)*, pp. 12316–12325, 2021. doi: 10.1109/CVPR46437.2021.01214.

Yaron Lipman, Ricky T. Q. Chen, Heli Ben-Hamu, Maximilian Nickel, and Matthew Le. Flow matching for generative modeling. In *The Eleventh International Conference on Learning Representations*, 2023. URL https://openreview.net/forum?id=PqvMRDCJT9t.

Ze Liu, Yutong Lin, Yue Cao, Han Hu, Yixuan Wei, Zheng Zhang, Stephen Lin, and Baining Guo. Swin transformer: Hierarchical vision transformer using shifted windows. In *2021 IEEE/CVF International Conference on Computer Vision (ICCV)*, pp. 9992–10002, 2021. doi: 10.1109/ICCV48922.2021.00986.

Yingzhou Lu, Lulu Chen, Yuanyuan Zhang, Minjie Shen, Huazheng Wang, Xiao Wang, Capucine van Rechem, Tianfan Fu, and Wenqi Wei. Machine learning for synthetic data generation: A review, 2025. URL https://arxiv.org/abs/2302.04062.

Yishay Mansour, Mehryar Mohri, and Afshin Rostamizadeh. Domain adaptation: Learning bounds and algorithms. In *Proceedings of The 22nd Annual Conference on Learning Theory (COLT 2009)*, Montréal, Canada, 2009. URL http://www.cs.nyu.edu/~mohri/postscript/nadap.pdf.

Lan-Cuong Nguyen, Quan Nguyen-Tri, Bang Tran Khanh, Dung D. Le, Long Tran-Thanh, and Khoat Than. Provably improving generalization of few-shot models with synthetic data. In Aarti Singh, Maryam Fazel, Daniel Hsu, Simon Lacoste-Julien, Felix Berkenkamp, Tegan Maharaj, Kiri Wagstaff, and Jerry Zhu (eds.), *Proceedings of the 42nd International Conference on Machine Learning*, volume 267 of *Proceedings of Machine Learning Research*, pp. 46178–46195. PMLR, 13–19 Jul 2025. URL https://proceedings.mlr.press/v267/nguyen25h.html.

Curtis Northcutt, Anish Athalye, and Jonas Mueller. Pervasive label errors in test sets destabilize machine learning benchmarks. In J. Vanschoren and S. Yeung (eds.), *Proceedings of the Neural Information Processing Systems Track on Datasets and Benchmarks*, volume 1, 2021. URL https://datasets-benchmarks-proceedings.neurips.cc/paper_files/paper/2021/file/f2217062e9a397a1dca429e7d70bc6ca-Paper-round1.pdf.

Yongming Rao, Wenliang Zhao, Zheng Zhu, Jie Zhou, and Jiwen Lu. Gfnet: Global filter networks for visual recognition. *IEEE Transactions on Pattern Analysis and Machine Intelligence*, 45(9):10960–10973, 2023. doi: 10.1109/TPAMI.2023.3263824.

Ievgen Redko, Amaury Habrard, and Marc Sebban. Theoretical analysis of domain adaptation with optimal transport. In *Machine Learning and Knowledge Discovery in Databases (ECML PKDD)*, pp. 737–753. Springer, 2017.

Danilo Rezende and Shakir Mohamed. Variational inference with normalizing flows. In Francis Bach and David Blei (eds.), *Proceedings of the 32nd International Conference on Machine Learning*, volume 37 of *Proceedings of Machine Learning Research*, pp. 1530–1538, Lille, France, 07–09 Jul 2015. PMLR. URL https://proceedings.mlr.press/v37/rezende15.html.

Stephan R. Richter, Vibhav Vineet, Stefan Roth, and Vladlen Koltun. Playing for data: Ground truth from computer games. In Bastian Leibe, Jiri Matas, Nicu Sebe, and Max Welling (eds.), *Computer Vision – ECCV 2016*, pp. 102–118, Cham, 2016. Springer International Publishing. ISBN 978-3-319-46475-6.

David E. Rumelhart, Geoffrey E. Hinton, and Ronald J. Williams. Learning representations by back-propagating errors. *Nature*, 323(6088):533–536, 1986. ISSN 1476-4687. doi: 10.1038/323533a0. URL https://doi.org/10.1038/323533a0.

Vignesh Sampath, Iñaki Maurtua, Juan José Aguilar Martín, and Aitor Gutierrez. A survey on generative adversarial networks for imbalance problems in computer vision tasks. *Journal of Big Data*, 8(1), 2021. ISSN 2196-1115. doi: 10.1186/s40537-021-00414-0. URL https://doi.org/10.1186/s40537-021-00414-0.

Mark Sandler, Andrew Howard, Menglong Zhu, Andrey Zhmoginov, and Liang-Chieh Chen. Mobilenetv2: Inverted residuals and linear bottlenecks. In *2018 IEEE/CVF Conference on Computer Vision and Pattern Recognition*, pp. 4510–4520, 2018. doi: 10.1109/CVPR.2018.00474.

Mert Bulent Sariyildiz, Karteek Alahari, Diane Larlus, and Yannis Kalantidis. Fake it till you make it: Learning transferable representations from synthetic imagenet clones. In *2023 IEEE/CVF Conference on Computer Vision and Pattern Recognition (CVPR)*, pp. 8011–8021, 2023. doi: 10.1109/CVPR52729.2023.00774.

Manuel Schwonberg, Joshua Niemeijer, Jan-Aike Termöhlen, Nico M Schmidt, Hanno Gottschalk, Tim Fingscheidt, et al. Survey on unsupervised domain adaptation for semantic segmentation for visual perception in automated driving. *IEEE Access*, 11:54296–54336, 2023.

Jian Shen, Yanru Qu, Weinan Zhang, and Yong Yu. Wasserstein distance guided representation learning for domain adaptation. In *Proceedings of the AAAI Conference on Artificial Intelligence*, volume 32, 2018.

Amitis Shidani, Tyler Farghly, Yang Sun, Habib Ganjgahi, and George Deligiannidis. Beyond real data: Synthetic data through the lens of regularization, 2026. URL https://arxiv.org/abs/2510.08095.

Hidetoshi Shimodaira. Improving predictive inference under covariate shift by weighting the log-likelihood function. *Journal of Statistical Planning and Inference*, 90(2):227–244, 2000. ISSN 0378-3758. doi: https://doi.org/10.1016/S0378-3758(00)00115-4. URL https://www.sciencedirect.com/science/article/pii/S0378375800001154.

Ilia Shumailov, Zakhar Shumaylov, Yiren Zhao, Nicolas Papernot, Ross Anderson, and Yarin Gal. Ai models collapse when trained on recursively generated data. *Nature*, 631:755–759, 2024. doi: 10.1038/s41586-024-07566-y. URL https://www.nature.com/articles/s41586-024-07566-y.

Christian Szegedy, Wei Liu, Yangqing Jia, Pierre Sermanet, Scott Reed, Dragomir Anguelov, Dumitru Erhan, Vincent Vanhoucke, and Andrew Rabinovich. Going deeper with convolutions. In *2015 IEEE Conference on Computer Vision and Pattern Recognition (CVPR)*, pp. 1–9, 2015. doi: 10.1109/CVPR.2015.7298594.

Ilya O Tolstikhin, Neil Houlsby, Alexander Kolesnikov, Lucas Beyer, Xiaohua Zhai, Thomas Unterthiner, Jessica Yung, Andreas Steiner, Daniel Keysers, Jakob Uszkoreit, Mario Lucic, and Alexey Dosovitskiy. Mlp-mixer: An all-mlp architecture for vision. In M. Ranzato, A. Beygelzimer, Y. Dauphin, P.S. Liang, and J. Wortman Vaughan (eds.), *Advances in Neural Information Processing Systems*, volume 34, pp. 24261–24272. Curran Associates, Inc., 2021. URL https://proceedings.neurips.cc/paper_files/paper/2021/file/cba0a4ee5ccd02fda0fe3f9a3e7b89fe-Paper.pdf.

Martin J. Wainwright. *High-Dimensional Statistics: A Non-Asymptotic Viewpoint*. Cambridge Series in Statistical and Probabilistic Mathematics. Cambridge University Press, Cambridge, 2019.

Saining Xie, Alexander Kirillov, Ross Girshick, and Kaiming He. Exploring randomly wired neural networks for image recognition. In *2019 IEEE/CVF International Conference on Computer Vision (ICCV)*, pp. 1284–1293, 2019. doi: 10.1109/ICCV.2019.00137.

Shirong Xu, Will Wei Sun, and Guang Cheng. Utility theory of synthetic data generation. *arXiv preprint arXiv:2305.10015*, 2023. URL https://arxiv.org/abs/2305.10015.

Fisher Yu and Vladlen Koltun. Multi-scale context aggregation by dilated convolutions, 2016. URL https://arxiv.org/abs/1511.07122.

Zihan Zhang, Lei Shi, and Ding-Xuan Zhou. Classification with deep neural networks and logistic loss. *Journal of Machine Learning Research*, 25(125):1–117, 2024. URL http://jmlr.org/papers/v25/22-0049.html.

# A Supplementary Proofs

## A.1 Proof of Theorem 1

*Proof of theorem 1.* We will write $R = R_Q$ for this proof. We will first bound the expected KL divergence. It can be rewritten in terms of the risk functional as

$$\begin{aligned}
\mathbb{E}_{Q_X}[\mathrm{KL}_X(p \parallel p_\theta)] &= \mathbb{E}_{Q_X}[H_X(p, p_\theta)] - \mathbb{E}_{Q_X}[H_X(p)] \\
&= \mathbb{E}_{Q_X}[-p(0|X)\log p_\theta(0|X) - p(1|X)\log p_\theta(1|X)] - \mathbb{E}_{Q_X}[H_X(p)] \\
&= \mathbb{E}_Q[\ell(\theta, X, Y)] - \mathbb{E}_{Q_X}[H_X(p)] \\
&= R(\theta) - \mathbb{E}_{Q_X}[H_X(p)] \\
&= \left[R(\theta^*) - \mathbb{E}_{Q_X}[H_X(p)]\right] + [R(\theta) - R(\theta^*)] \\
&= \mathbb{E}_{Q_X}[\mathrm{KL}_X(p \parallel p_{\theta^*})] + [R(\theta) - R(\theta^*)] \\
&= \underbrace{\varepsilon_{\mathcal{F}}}_{\geq 0} + [R(\theta) - R(\theta^*)]
\end{aligned}$$

It remains to bound $R\left(\hat{\theta}_n\right) - R(\theta^*)$. Consider the following decomposition:

$$R\left(\hat{\theta}_n\right) - R(\theta^*) = \underbrace{\left\{R\left(\hat{\theta}_n\right) - \hat{R}\left(\hat{\theta}_n, V_n\right)\right\}}_{T_1} + \underbrace{\left\{\hat{R}\left(\hat{\theta}_n, V_n\right) - \hat{R}(\theta^*, V_n)\right\}}_{T_2 \leq 0} + \underbrace{\left\{\hat{R}(\theta^*, V_n) - R(\theta^*)\right\}}_{T_3}.$$

Using Lemma 11 this will give us a bound of the form

$$\mathrm{err}_Q\left(\hat{y}_{\hat{\theta}_n}\right) - \mathrm{err}_Q(y^*) \lesssim \sqrt{(\varepsilon_{\mathcal{F}} + T_1 + T_3)}. \tag{1}$$

We have $T_1 = R\left(\hat{\theta}_n\right) - \hat{R}\left(\hat{\theta}_n, V_n\right) \leq \sup_{\theta \in \Theta}\left|R(\theta) - \hat{R}(\theta, V_n)\right| = \|Q_n - Q\|_{\mathcal{L}}$, where $Q_n := \frac{1}{n}\sum_{i=1}^n \delta_{(X_i, Y_i)}$ is the empirical measure of $V_n$ and $\|\mu\|_{\mathcal{L}} := \sup_{\theta \in \Theta}\left|\int \ell(\theta, \cdot)\,\mathrm{d}\mu\right|$. Since $\theta \mapsto \ell(\theta, x, y)$ is continuous and $\Theta$ is compact, all suprema over $\Theta$ in this proof coincide with suprema over a countable dense subset and are therefore measurable. By Theorem 4.10 of Wainwright (2019) applied to $\mathcal{L}$ with uniform bound $b = \log(1/\gamma)$, for any $\epsilon \geq 0$ we have

$$\|Q_n - Q\|_{\mathcal{L}} \leq 2\mathcal{R}_n(\mathcal{L}) + \epsilon \tag{2}$$

with $Q$-probability at least $1 - \exp\left(-\frac{n\epsilon^2}{2b^2}\right)$, where $\mathcal{R}_n(\mathcal{L})$ is the Rademacher complexity

$$\mathcal{R}_n(\mathcal{L}) = \mathbb{E}_{V_n, \varepsilon}\left[\sup_{\theta \in \Theta}\left|\frac{1}{n}\sum_{i=1}^n \varepsilon_i \ell(\theta, x_i, y_i)\right|\right],$$

and $\varepsilon_i$ are i.i.d. Rademacher variables independent of $V_n$. Setting $\exp\left(-\frac{n\epsilon^2}{2b^2}\right) = \delta/2$ and solving for $\epsilon$ gives $\epsilon = \log(1/\gamma)\sqrt{2\log(2/\delta)/n}$. To bound $\mathcal{R}_n(\mathcal{L})$, we apply Theorem 5.22 of Wainwright (2019), which establishes that for any zero-mean sub-Gaussian process $\{X_\theta\}_{\theta \in \mathbb{T}}$ with respect to a pseudometric $\rho_X$, and any $\alpha \in [0, D]$, it holds that:

$$\mathbb{E}\left[\sup_{\theta, \tilde{\theta} \in \mathbb{T}}(X_\theta - X_{\tilde{\theta}})\right] \leq 2\,\mathbb{E}\left[\sup_{\substack{\vartheta, \vartheta' \in \mathbb{T} \\ \rho_X(\vartheta, \vartheta') \leq \alpha}}(X_\vartheta - X_{\vartheta'})\right] + 32\,\mathcal{J}(\alpha/4; D), \tag{3}$$

where $D := \mathrm{diam}(\mathbb{T})$ is the $\rho_X$-diameter and $\alpha/4$ and $D$ denote the integration limits. The construction that follows is carried out for a fixed realisation of $V_n$, so that loss values are deterministic and only $\varepsilon$ is random. The expectation over $V_n$ is taken at the end. Define the process

$$Z_\theta := \frac{1}{\sqrt{n}}\sum_{i=1}^n \varepsilon_i \ell(\theta, x_i, y_i),$$

which is zero-mean since $\mathbb{E}[\varepsilon_i] = 0$. Define the empirical $L^2$-pseudometric on $\Theta$ by

$$\rho_X(\theta,\theta') := \|\ell_\theta - \ell_{\theta'}\|_{L^2(Q_n)} = \sqrt{\frac{1}{n}\sum_{i=1}^{n}\left(\ell(\theta,x_i,y_i) - \ell(\theta',x_i,y_i)\right)^2}.$$

This is only a pseudometric, since distinct parameters can induce identical losses on the sample. We verify that $\{Z_\theta\}$ is sub-Gaussian with respect to $\rho_X$. Writing $a_i := \ell(\theta,x_i,y_i) - \ell(\theta',x_i,y_i)$, the $a_i$ are deterministic for fixed $V_n$, so $\varepsilon_i a_i \in [-|a_i|, +|a_i|]$ is a deterministic bound and Hoeffding's lemma gives $\mathbb{E}_\varepsilon\left[e^{t\varepsilon_i a_i}\right] \le \exp(t^2 a_i^2/2)$. By independence of the $\varepsilon_i$,

$$\mathbb{E}_\varepsilon\left[e^{\lambda(Z_\theta - Z_{\theta'})}\right] = \prod_{i=1}^{n}\mathbb{E}_\varepsilon\left[e^{\frac{\lambda}{\sqrt{n}}\varepsilon_i a_i}\right] \le \prod_{i=1}^{n} e^{\frac{\lambda^2 a_i^2}{2n}} = e^{\frac{\lambda^2\rho_X^2(\theta,\theta')}{2}}.$$

Hence $\{Z_\theta : \theta \in \Theta\}$ is sub-Gaussian with respect to $\rho_X$ in the sense of Definition 5.16 of Wainwright (2019). The diameter satisfies $D := \sup_{\theta,\theta'\in\Theta}\rho_X(\theta,\theta') \le \sup_{\theta,\theta'}\|\ell_\theta - \ell_{\theta'}\|_\infty \le \log((1-\gamma)/\gamma)$. Since $\rho_X(\theta,\theta') \le \|\ell_\theta - \ell_{\theta'}\|_\infty$, every $t$-cover of $\mathcal{L}$ in $\|\cdot\|_\infty$ induces a $t$-cover of $\Theta$ in $\rho_X$, so the $\rho_X$-covering numbers are bounded by $N(t,\mathcal{L},\|\cdot\|_\infty)$. By Wainwright (2019) Theorem 5.22 with $\alpha = 0$, the pairwise supremum is therefore controlled by the Dudley integral from Section 3:

$$\mathbb{E}_\varepsilon\left[\sup_{\theta,\theta'\in\Theta}(Z_\theta - Z_{\theta'})\right] \le 32\,\mathcal{J}(\mathcal{L}).$$

The first term in the r.h.s. of equation 3 vanishes at $\alpha = 0$. Although $\rho_X$ is only a pseudometric, $\rho_X(\theta,\theta') = 0$ forces $\ell(\theta,x_i,y_i) = \ell(\theta',x_i,y_i)$ for every $i$, hence $Z_\theta - Z_{\theta'} = 0$ identically. The upper integration limit exceeds the true diameter $D \le \log((1-\gamma)/\gamma)$, but the integrand vanishes for radii larger than $D$. Theorem 5.22 bounds the pairwise supremum, whereas $\mathcal{R}_n(\mathcal{L})$ involves $\mathbb{E}[\sup_\theta|Z_\theta|]$. To pass between them, fix any $\theta_0 \in \Theta$ and write $S := \sup_{\theta',\theta''\in\Theta}(Z_{\theta'} - Z_{\theta''})$. For every $\theta$,

$$Z_\theta = Z_{\theta_0} + \left(Z_\theta - Z_{\theta_0}\right) \le |Z_{\theta_0}| + S \quad \text{and} \quad -Z_\theta = -Z_{\theta_0} + \left(Z_{\theta_0} - Z_\theta\right) \le |Z_{\theta_0}| + S,$$

hence $|Z_\theta| \le |Z_{\theta_0}| + S$, and therefore

$$\begin{aligned}
\mathcal{R}_n(\mathcal{L}) &= \frac{1}{\sqrt{n}}\,\mathbb{E}_{V_n,\varepsilon}\left[\sup_\theta|Z_\theta|\right] \\
&\le \frac{1}{\sqrt{n}}\,\mathbb{E}_{V_n,\varepsilon}\left[|Z_{\theta_0}| + \sup_{\theta,\theta'}(Z_\theta - Z_{\theta'})\right] && \text{(triangle ineq., see below)} \\
&= \frac{1}{\sqrt{n}}\left(\mathbb{E}_{V_n,\varepsilon}\,|Z_{\theta_0}| + \mathbb{E}_{V_n,\varepsilon}\left[\sup_{\theta,\theta'}(Z_\theta - Z_{\theta'})\right]\right) && \text{(linearity)} \\
&\le \frac{1}{\sqrt{n}}\left(\mathbb{E}_{V_n}\left[\left(\mathbb{E}_\varepsilon Z_{\theta_0}^2\right)^{1/2}\right] + 32\,\mathcal{J}(\mathcal{L})\right) && \text{(Jensen; Thm 5.22 at } \alpha = 0) \\
&= \frac{1}{\sqrt{n}}\left(\mathbb{E}_{V_n}\left[\left\|\ell_{\theta_0}\right\|_{L^2(Q_n)}\right] + 32\,\mathcal{J}(\mathcal{L})\right) && \left(\mathbb{E}_\varepsilon\left[\varepsilon_i\varepsilon_j\right] = \delta_{ij}\right) \\
&\le \frac{1}{\sqrt{n}}\left(\left\|\ell_{\theta_0}\right\|_\infty + 32\,\mathcal{J}(\mathcal{L})\right) && \left(\|\cdot\|_{L^2(Q_n)} \le \|\cdot\|_\infty\right) \\
&\le \frac{b + 32\,\mathcal{J}(\mathcal{L})}{\sqrt{n}}, && \left(\left\|\ell_{\theta_0}\right\|_\infty \le b\right)
\end{aligned}$$

where the fourth line uses $\mathbb{E}_\varepsilon\,|Z_{\theta_0}| \le \left(\mathbb{E}_\varepsilon Z_{\theta_0}^2\right)^{1/2} = \left\|\ell_{\theta_0}\right\|_{L^2(Q_n)} \le \left\|\ell_{\theta_0}\right\|_\infty$ together with the Theorem 5.22 bound above, and the last uses $\left\|\ell_{\theta_0}\right\|_\infty \le b$. Therefore with probability at least $1 - \delta/2$, (2) yields that $T_1$ is bounded by

$$T_1 \le \frac{2b + 64\,\mathcal{J}(\mathcal{L})}{\sqrt{n}} + \log\left(\frac{1}{\gamma}\right)\sqrt{\frac{2\log(2/\delta)}{n}}.$$

We will now bound $T_3$. Since $\theta^*$ is non-random, $T_3 = \hat{R}(\theta^*, V_n) - R(\theta^*) = \frac{1}{n}\sum_{i=1}^n \ell(\theta^*, X_i, Y_i) - \mathbb{E}[\ell(\theta^*, X, Y)]$ is an average of i.i.d. centered random variables, each bounded in $[\log(1/(1-\gamma)), \log(1/\gamma)]$ with range $\log((1-\gamma)/\gamma)$. By Hoeffding's inequality, $Q^{\otimes n}(T_3 \geq t) \leq \exp\{-2nt^2/\log^2((1-\gamma)/\gamma)\}$. Setting $\exp\{-2nt^2/\log^2((1-\gamma)/\gamma)\} = \delta/2$ and solving for $t$ gives $t = \log((1-\gamma)/\gamma)\sqrt{\log(2/\delta)/2n}$. Therefore with probability at least $1-\delta/2$,

$$T_3 \leq \log\left(\frac{1-\gamma}{\gamma}\right)\sqrt{\frac{\log(2/\delta)}{2n}}.$$

By a union bound, both bounds hold simultaneously with probability at least $1-\delta$ over $V_n \sim Q^{\otimes n}$. Since $T_2 \leq 0$, we have $R(\hat{\theta}_n) - R(\theta^*) \leq T_1 + T_3$, hence

$$\begin{aligned}
R(\hat{\theta}_n) - R(\theta^*) &\leq \frac{2b + 64\,\mathcal{J}(\mathcal{L})}{\sqrt{n}} + \log\left(\frac{1}{\gamma}\right)\sqrt{\frac{2\log(2/\delta)}{n}} + \log\left(\frac{1-\gamma}{\gamma}\right)\sqrt{\frac{\log(2/\delta)}{2n}} \\
&\leq \frac{2\log(1/\gamma)}{\sqrt{n}} + \frac{64\,\mathcal{J}(\mathcal{L})}{\sqrt{n}} + \sqrt{2}\log\left(\frac{1}{\gamma}\right)\frac{\sqrt{\log(2/\delta)}}{\sqrt{n}} + \frac{1}{\sqrt{2}}\log\left(\frac{1}{\gamma}\right)\frac{\sqrt{\log(2/\delta)}}{\sqrt{n}} \\
&\lesssim \frac{\mathcal{J}(\mathcal{L})}{\sqrt{n}} + \frac{\log(1/\gamma)}{\sqrt{n}} + \log\left(\frac{1}{\gamma}\right)\frac{\sqrt{\log(2/\delta)}}{\sqrt{n}} \\
&= \frac{\mathcal{J}(\mathcal{L}) + \log(1/\gamma)\left(1 + \sqrt{\log(2/\delta)}\right)}{\sqrt{n}},
\end{aligned}$$

Together with the decomposition $\mathbb{E}_{Q_X}\left[\mathrm{KL}_X\left(p \parallel p_{\hat{\theta}_n}\right)\right] = \varepsilon_{\mathcal{F}} + \left(R(\hat{\theta}_n) - R(\theta^*)\right)$ from the beginning of the proof, and applying Lemma 11 then gives the result. In both bounds, $\lesssim$ hides only universal constants, independent of $n$, $\delta$, and $\gamma$. ∎

## A.2 Concrete rate for fully-connected truncated deep ReLU networks

**Remark 9** (Explicit rate)**.** Assume ($\kappa$) and $f^* \coloneqq \mathtt{logit} \circ \eta$. Choose the bound of the model class as $B \coloneqq F = \log((1-\kappa)/\kappa)$, so that $\gamma = \sigma(-B) = \kappa$. Let $g_\theta : \mathbb{R}^d \to \mathbb{R}$ be a fully connected ReLU network, cf. Equation (5) of depth $L$ and hidden width $N \coloneqq d+2$, with parameters $\theta$ ranging over $\Theta \coloneqq [-W, W]^m$ for a weight bound $W \geq 2$ and $m$ the number of parameters, and let $\mathcal{F} \coloneqq \{T_B \circ g_\theta : \theta \in \Theta\}$ and $T_B(t) \coloneqq \max\{-B, \min\{B, t\}\}$ truncation with $B$. Then $\mathcal{F}$ is a model class in the sense of Section 3 with bound $B$, and

$$\varepsilon_{\mathcal{F}} \leq \frac{\inf_{f\in\mathcal{F}} \|f^* - f\|_\infty^2}{16\,\kappa(1-\kappa)}, \qquad \mathcal{J}(\mathcal{L}) \lesssim L\sqrt{\log W}.$$

*Proof.* Each $f_\theta = T_B \circ g_\theta$ is continuous with $\|f_\theta\|_\infty \leq B$, so $\mathcal{F}$ is a model class with bound $B$, and $\gamma = \sigma(-B) = \kappa$. By ($\kappa$), $f^* = \mathtt{logit} \circ \eta$ satisfies $\|f^*\|_\infty \leq B$. Throughout, $r_{\mathcal{X}} \coloneqq \sup_{x\in\mathcal{X}} \|x\|_\infty \vee 1 < \infty$ by compactness of $\mathcal{X}$, and $C, C', \ldots$ denote constants depending only on $d$ and $r_{\mathcal{X}}$.

*Step 1: the approximation floor.* For $\mu \ll \nu$ on a discrete space, $\log t \leq t - 1$ with $t = \mu(y)/\nu(y)$ gives

$$
\begin{aligned}
\mathrm{KL}(\mu \parallel \nu) &= \sum_y \mu(y) \log \frac{\mu(y)}{\nu(y)} \\
&\le \sum_y \mu(y) \left( \frac{\mu(y)}{\nu(y)} - 1 \right) \\
&= \sum_y \frac{\mu(y)^2}{\nu(y)} - \sum_y \mu(y) \\
&= \sum_y \frac{\mu(y)^2}{\nu(y)} - 2 \sum_y \mu(y) + \sum_y \nu(y) \\
&= \sum_y \frac{\mu(y)^2 - 2\mu(y)\nu(y) + \nu(y)^2}{\nu(y)} \\
&= \sum_y \frac{(\mu(y) - \nu(y))^2}{\nu(y)} = \chi^2(\mu \parallel \nu).
\end{aligned}
$$

Writing $\eta_f := \sigma \circ f$, which takes values in $[\kappa, 1-\kappa]$ whenever $\|f\|_\infty \le B$,

$$
\begin{aligned}
\chi^2\big(\mathrm{Bern}_{\eta(x)} \| \,\mathrm{Bern}_{\eta_f(x)}\big) &= \frac{\big(\eta(x) - \eta_f(x)\big)^2}{\eta_f(x)} + \frac{\big((1-\eta(x)) - \big(1-\eta_f(x)\big)\big)^2}{1-\eta_f(x)} \\
&= \big(\eta(x) - \eta_f(x)\big)^2 \left( \frac{1}{\eta_f(x)} + \frac{1}{1-\eta_f(x)} \right) \\
&= \big(\eta(x) - \eta_f(x)\big)^2 \frac{\big(1-\eta_f(x)\big) + \eta_f(x)}{\eta_f(x)\big(1-\eta_f(x)\big)} \\
&= \frac{\big(\eta(x) - \eta_f(x)\big)^2}{\eta_f(x)\big(1-\eta_f(x)\big)} \\
&\le \frac{\big(\eta(x) - \eta_f(x)\big)^2}{\kappa(1-\kappa)}.
\end{aligned}
$$

Since $\sup_t \sigma'(t) = \sup_t \sigma(t)(1-\sigma(t)) \le \frac{1}{4}$, we have $|\eta(x) - \eta_f(x)| \le \frac{1}{4}\|f^* - f\|_\infty$, hence for every $x \in \mathcal{X}$

$$
\mathrm{KL}_x\big(p \parallel p_f\big) \le \frac{\|f^* - f\|_\infty^2}{16\,\kappa(1-\kappa)}. \tag{4}
$$

Since $R_Q(\theta) = \mathbb{E}_{Q_X}[H_X(p)] + \mathbb{E}_{Q_X}[\mathrm{KL}_X(p \parallel p_\theta)]$ and the first term does not depend on $\theta$, $\theta^*$ also minimizes $\mathbb{E}_{Q_X}[\mathrm{KL}_X(p \parallel p_\theta)]$. Hence $\varepsilon_{\mathcal{F}} \le \mathbb{E}_{Q_X}[\mathrm{KL}_X(p \parallel p_\theta)]$ for all $\theta \in \Theta$, and taking the infimum over $f \in \mathcal{F}$ in equation 4 gives the stated bound.

*Step 2: the entropy integral.* Write

$$
g_\theta = A_\theta^{(L+1)} \circ \mathrm{ReLU} \circ \cdots \circ \mathrm{ReLU} \circ A_\theta^{(1)} \tag{5}
$$

with $A_\theta^{(l)}(z) = W^{(l)} z + b^{(l)}$, and for $f \in \mathcal{F}$ let $\ell_f(x,y) := -\log p_f(y|x)$ with $p_f(y|x) := y\,\eta_f(x) + (1-y)\big(1 - \eta_f(x)\big)$. Both loss maps $t \mapsto -\log\sigma(t)$ and $t \mapsto -\log(1-\sigma(t))$ have derivatives bounded by 1 in absolute value, so $|\ell_f(x,y) - \ell_{f'}(x,y)| \le \|f - f'\|_\infty$

$$
\begin{aligned}
|\ell_f(x,y) - \ell_{f'}(x,y)| &= \big|\log p_{f'}(y|x) - \log p_f(y|x)\big| \\
&= \begin{cases} |\log\sigma(f'(x)) - \log\sigma(f(x))|, & y = 1 \\ |\log(1-\sigma(f'(x))) - \log(1-\sigma(f(x)))|, & y = 0 \end{cases} \\
&\le |f(x) - f'(x)| \le \|f - f'\|_\infty,
\end{aligned}
$$

and therefore $N(t, \mathcal{L}, \|\cdot\|_\infty) \le N(t, \mathcal{F}, \|\cdot\|_\infty)$.

The hidden representations are uniformly bounded. With $h^{(0)} := x$ and $h^{(l)} := \mathrm{ReLU}\Big(A_\theta^{(l)}\big(h^{(l-1)}\big)\Big)$, every row of $W^{(l)}$ has at most $N$ entries, all parameters are bounded by $W$, and ReLU does not increase absolute values, so $\|h^{(l)}\|_\infty \le NW\|h^{(l-1)}\|_\infty + W$. Using $W \le NWr_{\mathcal{X}}$ (as $N, r_{\mathcal{X}} \ge 1$), induction over $l$ gives

$$\|h^{(l)}\|_\infty \le (2NW)^l r_{\mathcal{X}} =: M_l, \qquad l = 0, ..., L.$$

Next, $\theta \mapsto f_\theta$ is Lipschitz. Let $\|\theta - \theta'\|_\infty \le \tau$ and interpolate through hybrids $\theta = \theta^{(0)}, ..., \theta^{(L+1)} = \theta'$, where $\theta^{(k)}$ carries the parameters of $\theta'$ in layers $1, ..., k$ and those of $\theta$ elsewhere; all hybrids lie in $\Theta$, so the bounds $M_l$ apply. Consecutive hybrids $\theta^{(l-1)}$ and $\theta^{(l)}$ differ only in layer $l$, where they use the parameters $\big(W^{(l)}, b^{(l)}\big)$ of $\theta$ and $\big(W'^{(l)}, b'^{(l)}\big)$ of $\theta'$, respectively. Both apply layer $l$ to the same input $h := h^{(l-1)}$ with $\|h\|_\infty \le M_{l-1}$. Since all entries differ by at most $\tau$ and each row of $W^{(l)}$ has at most $N$ entries, for every output coordinate $i$,

$$\begin{aligned}\left|\big(W'^{(l)}h + b'^{(l)}\big)_i - \big(W^{(l)}h + b^{(l)}\big)_i\right| &= \left|\sum_j \Big(W_{ij}'^{(l)} - W_{ij}^{(l)}\Big)h_j + \Big(b_i'^{(l)} - b_i^{(l)}\Big)\right| \\ &\le \sum_j \left|W_{ij}'^{(l)} - W_{ij}^{(l)}\right| |h_j| + \left|b_i'^{(l)} - b_i^{(l)}\right| \\ &\le N\tau M_{l-1} + \tau = (NM_{l-1} + 1)\tau.\end{aligned}$$

Hence the outputs of layer $l$ differ by at most $(NM_{l-1} + 1)\tau$ in $\|\cdot\|_\infty$. Each of the $L + 1 - l$ subsequent affine maps multiplies this difference by at most $NW$, while ReLU and $T_B$ are 1-Lipschitz and do not increase it. Summing over the $L + 1$ layers and using $NM_{l-1} + 1 \le 2N(2NW)^{l-1}r_{\mathcal{X}}$ gives

$$\begin{aligned}\|f_\theta - f_{\theta'}\|_\infty &\le \sum_{l=1}^{L+1} \|f_{\theta^{(l-1)}} - f_{\theta^{(l)}}\|_\infty \\ &\le \sum_{l=1}^{L+1} (NW)^{L+1-l}(NM_{l-1} + 1)\tau \\ &\le \sum_{l=1}^{L+1} (NW)^{L+1-l}\, 2NM_{l-1}\, \tau && (NM_{l-1} \ge 1) \\ &= \sum_{l=1}^{L+1} (NW)^{L+1-l}\, 2N(2NW)^{l-1} r_{\mathcal{X}}\, \tau && (\text{definition of } M_{l-1}) \\ &\le \sum_{l=1}^{L+1} (2NW)^{L+1-l}\, 2Nr_{\mathcal{X}}\, (2NW)^{l-1}\, \tau \\ &= \sum_{l=1}^{L+1} 2Nr_{\mathcal{X}}\, (2NW)^L\, \tau \\ &= (L+1)\, 2Nr_{\mathcal{X}}\, (2NW)^L\, \tau \\ &\le L\, 4Nr_{\mathcal{X}}\, (2NW)^L\, \tau && (L + 1 \le 2L) \\ &= L\, C\, (2NW)^L\, \tau && (C := 4Nr_{\mathcal{X}}) \\ &\le L\, C\, (CW)^L\, \tau && (2N \le C \text{ as } r_{\mathcal{X}} \ge 1) \\ &\le L\, CW\, (CW)^L\, \tau && (W \ge 1) \\ &= \underbrace{L\, (CW)^{L+1}}_{=:\Lambda}\, \tau. \end{aligned} \tag{6}$$

By Wainwright (2019, Example 5.8), $\log N(\delta; [-1, 1]^m, \|\cdot\|_\infty) \le m \log(1 + 2/\delta)$. Since $\Theta = W[-1, 1]^m$, taking $\delta = t/(W\Lambda)$ gives

$$N(t/\Lambda; \Theta, \|\cdot\|_\infty) \le \left(1 + \frac{2W\Lambda}{t}\right)^m,$$

and by the Lipschitz bound the image of such a cover under $\theta \mapsto f_\theta$ is a $t$-cover of $\mathcal{F}$, because every $\theta \in \Theta$ has a cover element $\theta_i$ with $\|\theta - \theta_i\|_\infty \le t/\Lambda$, so that equation 6 gives $\left\|f_\theta - f_{\theta_i}\right\|_\infty \le \Lambda \cdot t/\Lambda = t$.

Since each of the $L+1$ affine maps has at most $N(N+1)$ parameters,

$$m \le N(N+1)(L+1) \le 2N(N+1)\,L = \underbrace{2N(N+1)}_{=:C_1}L \qquad (L+1 \le 2L).$$

Moreover,

$$\begin{aligned}
\log(1+2W\Lambda) &\le \log(3W\Lambda) && (W\Lambda \ge 1)\\
&= \log 3 + \log W + \log L + (L+1)\log C + (L+1)\log W && (\Lambda = L(CW)^{L+1})\\
&\le \log 3 + \log W + L + 2L\log C + 2L\log W && (\log L \le L,\, L+1 \le 2L)\\
&\le L\log 3 + L\log W + L + 2L\log C + 2L\log W && (L \ge 1)\\
&= (\log 3 + 1 + 2\log C)\,L + 3L\log W\\
&\le (\log 3 + 1 + 2\log C)\,\frac{L\log W}{\log 2} + 3L\log W && (\log W \ge \log 2)\\
&= \underbrace{\left(\frac{\log 3 + 1 + 2\log C}{\log 2} + 3\right)}_{=:C_2} L\log W.
\end{aligned}$$

Hence

$$\begin{aligned}
\log N(t, \mathcal{F}, \|\cdot\|_\infty) &\le m\log(1+2W\Lambda/t) && \text{(covering bound)}\\
&\le m(\log(1+2W\Lambda) + (\log(1/t) \vee 0)) && (\log(1+a/t) \le \log(1+a) + (\log(1/t) \vee 0))\\
&\le C_1 L(C_2 L\log W + (\log(1/t) \vee 0)) && (\text{bounds on } m \text{ and } \log(1+2W\Lambda))\\
&= C_1C_2\,L^2\log W + C_1 L\,(\log(1/t) \vee 0)\\
&\le \underbrace{C_1\max\{C_2, 1\}}_{=:C_3}(L^2\log W + L(\log(1/t) \vee 0)).
\end{aligned}$$

Finally, since $\gamma = \kappa$, the upper integration limit in $\mathcal{J}(\mathcal{L})$ is $B$, and

$$
\begin{aligned}
\mathcal{J}(\mathcal{L}) &= \int_0^B \sqrt{\log N(t, \mathcal{L}, \|\cdot\|_\infty)}\,\mathrm{d}t \\
&\le \int_0^B \sqrt{\log N(t, \mathcal{F}, \|\cdot\|_\infty)}\,\mathrm{d}t && (N(t,\mathcal{L}) \le N(t,\mathcal{F})) \\
&\le \int_0^B \sqrt{C_3(L^2 \log W + L(\log(1/t) \vee 0))}\,\mathrm{d}t \\
&= \sqrt{C_3} \int_0^B \sqrt{L^2 \log W + L(\log(1/t) \vee 0)}\,\mathrm{d}t \\
&\le \sqrt{C_3} \int_0^B \Big(L\sqrt{\log W} + \sqrt{L}\,\sqrt{\log(1/t) \vee 0}\Big)\mathrm{d}t && \Big(\sqrt{a+b} \le \sqrt{a} + \sqrt{b}\Big) \\
&= \sqrt{C_3}\left(B\,L\sqrt{\log W} + \sqrt{L}\int_0^B \sqrt{\log(1/t) \vee 0}\,\mathrm{d}t\right) \\
&\le \sqrt{C_3}\left(B\,L\sqrt{\log W} + \sqrt{L}\int_0^1 \sqrt{\log(1/t)}\,\mathrm{d}t\right) && (\text{integrand vanishes for } t \ge 1) \\
&= \sqrt{C_3}\left(B\,L\sqrt{\log W} + \frac{\sqrt{\pi}}{2}\sqrt{L}\right) \\
&\le \sqrt{C_3}\left(B + \frac{\sqrt{\pi}}{2\sqrt{\log 2}}\right)L\sqrt{\log W} && \Big(\sqrt{L} \le L,\, 1 \le \sqrt{\log W/\log 2}\Big) \\
&= O\big(L\sqrt{\log W}\big).
\end{aligned}
$$

□

### A.3 Proof of Theorem 3

*Proof.* (i) By the change-of-measure formula, we have

$$
\mathbb{E}_{P_X}\Big[\mathrm{KL}_X\big(p \parallel p_{\hat{\theta}_n}\big)\Big] = \mathbb{E}_{Q_X}\left[\mathrm{KL}_X\big(p \parallel p_{\hat{\theta}_n}\big)\frac{\mathrm{d}P_X}{\mathrm{d}Q_X}\right] \le \left\|\frac{\mathrm{d}P_X}{\mathrm{d}Q_X}\right\|_{L^\infty(\mathcal{X}, Q_X)} \mathbb{E}_{X\sim Q_X}\Big[\mathrm{KL}_X\big(p \parallel p_{\hat{\theta}_n}\big)\Big].
$$

Application of Theorem 1 yields the assertion.

(ii) For fixed but arbitrary $x \in \mathcal{X}$, we have $\mathrm{KL}_x\big(p|\ |p_{\hat{\theta}_n}\big) \le \left|p(0|x)\log\frac{p(0|x)}{p_{\hat{\theta}_n}(0|x)} + p(1|x)\log\frac{p(1|x)}{p_{\hat{\theta}_n}(1|x)}\right| \le \frac{2}{e} + H_x\big(p|\ |p_{\hat{\theta}_n}\big) < \infty$, because $\max_{t\in[0,1]}|t\log t| = 1/e$ and the cross-entropy term is bounded by $\log(1/\gamma)$, since $\|f_\theta\|_\infty \le B$. Now,

$$
\begin{aligned}
\mathbb{E}_{P_X}\Big[\mathrm{KL}_X\big(p|\ |p_{\hat{\theta}_n}\big)\Big] &\le \mathbb{E}_{Q_X}\left[\left|\frac{\mathrm{d}P_X}{\mathrm{d}Q_X}\mathrm{KL}_X\big(p|\ |p_{\hat{\theta}_n}\big)\right|\right] = \left\|\frac{\mathrm{d}P_X}{\mathrm{d}Q_X}\mathrm{KL}_X\big(p|\ |p_{\hat{\theta}_n}\big)\right\|_{\mathrm{L}^1(Q_X)} \\
&\le \left\|\frac{\mathrm{d}P_X}{\mathrm{d}Q_X}\right\|_{\mathrm{L}^s(Q_X)} \left\|\mathrm{KL}_X\big(p|\ |p_{\hat{\theta}_n}\big)\right\|_{\mathrm{L}^t(Q_X)} = \left\|\frac{\mathrm{d}P_X}{\mathrm{d}Q_X}\right\|_{\mathrm{L}^s(Q_X)} \mathbb{E}_{Q_X}\Big[\mathrm{KL}_X^t\big(p|\ |p_{\hat{\theta}_n}\big)\Big]^{1/t},
\end{aligned}
$$

where for $f_n \le C$, $\mathbb{E}^{1/t}[\![|f_n|^t]\!] = \mathbb{E}^{1/t}[\![|f_n|^{1+(t-1)}]\!] = \mathbb{E}^{1/t}[\![|f_n|\,|f_n|^{t-1}]\!] \le C^{(t-1)/t}\mathbb{E}^{1/t}[\![f_n]\!]$. Now the assertion follows from an application of Theorem 1 and Lemma 11. □

### A.4 Proof of Theorem 4

*Proof.* Note that $\tilde{y}$ is measurable. Write $D := \{x \in \mathcal{X} : \hat{y}_\theta(x) \ne \tilde{y}(x)\}$.

Let $\nu$ be a finite measure on $\mathcal{X} \times \mathcal{Y}$ with covariate marginal $\nu_X$. The events $\{\hat{y}_\theta(X) \ne Y\}$ and $\{\tilde{y}(X) \ne Y\}$ coincide on $D^c \times \mathcal{Y}$, so their symmetric difference is contained in $D \times \mathcal{Y}$. Since $\nu(E) \le \nu(E') + \nu(E \setminus E')$ for any measurable $E, E'$, and symmetrically,

$$|\mathrm{err}_\nu(\hat{y}_\theta) - \mathrm{err}_\nu(\tilde{y})| \leq \nu(D \times \mathcal{Y}) = \nu_X(D). \tag{7}$$

As $P^a_X$ and $P^s_X$ are restrictions of $P_X$, we have $\nu_X(D) \leq \varepsilon$ for $\nu \in \{Q, P^a, P^s\}$.

By construction $\tilde{y}$ and $y^*$ differ at most on $M_s$, and $Q_X(M_s) = 0$, so $\tilde{y} = y^*$ holds $Q_X$-a.e. Since $\mathrm{err}_Q$ depends on a classifier only through its values $Q_X$-a.e., this gives $\mathrm{err}_Q(\tilde{y}) = \mathrm{err}_Q(y^*)$. Combining with equation 7 for $\nu = Q$ and $Q_X(D) \leq \varepsilon$,

$$\mathrm{err}_Q(\hat{y}_\theta) \leq \mathrm{err}_Q(\tilde{y}) + Q_X(D) \leq \mathrm{err}_Q(y^*) + \varepsilon,$$

which is the first claim.

For every measurable $h$, using $\mathcal{Y} = \{0, 1\}$,

$$\mathrm{err}_{P^s}(h) = \int_{M_s} \sum_{y \in \mathcal{Y}} \mathbf{1}\{h(x) \neq y\}\, p(y|x)\, P_X(\mathrm{d}x) = \int_{M_s} p(1 - h(x) \,|\, x)\, P_X(\mathrm{d}x).$$

The integrand is at most $\max_y p(y|x)$ for every $x$, so $\mathrm{err}_{P^s}(h) \leq \int_{M_s} \max_y p(y|x)\, P_X(\mathrm{d}x)$ for every $h$. Since $\tilde{y} = \hat{y}_{\mathrm{adv}}$ on $M_s$, the integrand equals $\max_y p(y|x)$ for $h = \tilde{y}$, so this upper bound is attained and $\mathrm{err}_{P^s}(\tilde{y}) = \max_h \mathrm{err}_{P^s}(h)$. Equation 7 applied to $\nu = P^s$ gives $|\mathrm{err}_{P^s}(\hat{y}_\theta) - \mathrm{err}_{P^s}(\tilde{y})| \leq P^s_X(D)$, and dropping the absolute value yields $\mathrm{err}_{P^s}(\hat{y}_\theta) \geq \mathrm{err}_{P^s}(\tilde{y}) - P^s_X(D)$. Since $P^s_X$ is a restriction of $P_X$, we have $P^s_X(D) \leq P_X(D) \leq \varepsilon$, and therefore

$$\mathrm{err}_{P^s}(\hat{y}_\theta) \geq \mathrm{err}_{P^s}(\tilde{y}) - \varepsilon = \max_h \mathrm{err}_{P^s}(h) - \varepsilon,$$

which is the second claim. □

### A.5 Proof of Lemma 10

**Lemma 10.** *Assume* ($\kappa$), ($L_p$) *and* ($L_f$)*. Then, for every* $\theta \in \Theta$*,*

$$\mathbb{E}_{X \sim P_X}[\mathrm{KL}_X(p \,\|\, p_\theta)] \leq \mathbb{E}_{X \sim Q_X}[\mathrm{KL}_X(p \,\|\, p_\theta)] + \big[L_f + (B + F)L_p\big] W_1(P_X, Q_X).$$

*Proof of Lemma 10.* Set $\eta(x) := p(1|x)$ and define $G(a, t) := \mathrm{KL}\big(\mathrm{Bern}_a \,\big\|\, \mathrm{Bern}_{\sigma(t)}\big)$. Using $\log \sigma(t) = t - \log(1 + e^t)$ and $\log(1 - \sigma(t)) = -\log(1 + e^t)$, we obtain

$$\begin{aligned}
G(a, t) &:= \mathrm{KL}\big(\mathrm{Bern}_a \,\big\|\, \mathrm{Bern}_{\sigma(t)}\big) \\
&= a \log\left(\frac{a}{\sigma(t)}\right) + (1 - a) \log\left(\frac{1 - a}{1 - \sigma(t)}\right) \\
&= a \log a - a \log \sigma(t) + (1 - a) \log(1 - a) - (1 - a) \log(1 - \sigma(t)) \\
&= a \log a - a(t - \log(1 + e^t)) + (1 - a) \log(1 - a) - (1 - a)(-\log(1 + e^t)) \\
&= a \log a - at + a \log(1 + e^t) + (1 - a) \log(1 - a) + (1 - a) \log(1 + e^t) \\
&= a \log a + (1 - a) \log(1 - a) + \log(1 + e^t) - at.
\end{aligned}$$

Its partial derivatives are

$$\partial_t G(a, t) = \sigma(t) - a, \qquad \partial_a G(a, t) = \log\left(\frac{a}{1 - a}\right) - t.$$

By ($\kappa$), $\eta(x) \in [\kappa, 1 - \kappa]$, and $|f_\theta(x)| \leq B$ by definition of the model class. Therefore, with $\sigma(t), a \leq 1$ and $t = f_\theta(x) \leq B$,

$$|\partial_t G(a, t)| \leq 1, \qquad |\partial_a G(a, t)| \leq F + B.$$

For any $x, x' \in \mathcal{X}$, the mean value theorem gives

$$\begin{aligned}
&|\mathrm{KL}_x(p \,\|\, p_\theta) - \mathrm{KL}_{x'}(p \,\|\, p_\theta)| \\
&\quad = |G(\eta(x), f_\theta(x)) - G(\eta(x'), f_\theta(x'))| \\
&\quad \leq |f_\theta(x) - f_\theta(x')| + (B + F)|\eta(x) - \eta(x')| \\
&\quad \leq \big[L_f + (B + F)L_p\big] d(x, x').
\end{aligned}$$

Hence $x \mapsto \mathrm{KL}_x(p \parallel p_\theta)$ is Lipschitz with constant at most $L_f + (B+F)L_p$. Set

$$L := L_f + (B+F)L_p$$

and let $\pi \in \Gamma(P_X, Q_X)$ be any coupling. Since the marginals of $\pi$ are $P_X$ and $Q_X$, respectively,

$$\begin{aligned}
&\left|\mathbb{E}_{X\sim P_X}[\mathrm{KL}_X(p \parallel p_\theta)] - \mathbb{E}_{X\sim Q_X}[\mathrm{KL}_X(p \parallel p_\theta)]\right| \\
&= \left|\int_{\mathcal{X}\times\mathcal{X}} (\mathrm{KL}_x(p \parallel p_\theta) - \mathrm{KL}_{x'}(p \parallel p_\theta))\,\mathrm{d}\pi(x,x')\right| \\
&\leq \int_{\mathcal{X}\times\mathcal{X}} |\mathrm{KL}_x(p \parallel p_\theta) - \mathrm{KL}_{x'}(p \parallel p_\theta)|\,\mathrm{d}\pi(x,x') \\
&\leq L \int_{\mathcal{X}\times\mathcal{X}} d(x,x')\,\mathrm{d}\pi(x,x').
\end{aligned}$$

Taking the infimum over all $\pi \in \Gamma(P_X, Q_X)$ yields

$$\left|\mathbb{E}_{X\sim P_X}[\mathrm{KL}_X(p \parallel p_\theta)] - \mathbb{E}_{X\sim Q_X}[\mathrm{KL}_X(p \parallel p_\theta)]\right| \leq \left[L_f + (B+F)L_p\right] W_1(P_X, Q_X).$$

The claimed one-sided inequality follows. □

**Lemma 11.** *Let $\nu$ be a probability distribution on $\mathcal{X}\times\mathcal{Y}$ with conditional $p$ and covariate marginal $\nu_X$. For any $\theta \in \Theta$ and the Bayes classifier $y^*$,* $\mathrm{err}_\nu(\hat{y}_\theta) - \mathrm{err}_\nu(y^*) \leq \sqrt{2\mathbb{E}_{\nu_X}[\mathrm{KL}_X(p \parallel p_\theta)]}$.

*Proof.* By Theorem 2.2 of Devroye et al. (1996), it holds that $\mathrm{err}_\nu(\hat{y}_\theta) - \mathrm{err}_\nu(y^*) \leq 2\mathbb{E}_{\nu_X}|\eta(X) - \eta_\theta(X)|$. For two Bernoulli distributions we have $\mathrm{TV}\left(\mathrm{Bern}_{\eta(x)}, \mathrm{Bern}_{\eta_\theta(x)}\right) = |\eta(x) - \eta_\theta(x)|$. By taking the expectation and applying Pinsker's and Jensen's inequalities, we obtain the result. □

### A.6 Proof of Theorem 7

*Proof.* By ($\kappa$), $f^*$ is bounded. Let $x \in U_\theta$. By definition of $U_\theta$ as a union of balls, there is $x' \in S$ with $x \in B(x', r_\theta(x'))$, i.e.

$$\|x - x'\| < r_\theta(x') = \min\left\{\frac{|f_\theta(x')|}{L_f}, \frac{|f^*(x')|}{L^*}\right\}.$$

Since $x' \in S$, we have $\hat{y}_\theta(x') = y^*(x')$. By ($L_f$), $|f_\theta(x) - f_\theta(x')| \leq L_f\|x - x'\| < |f_\theta(x')|$, so $f_\theta(x)$ has the same strict sign as $f_\theta(x')$ and hence $\hat{y}_\theta(x) = \hat{y}_\theta(x')$. By ($L^*$), $|f^*(x) - f^*(x')| \leq L^*\|x - x'\| < |f^*(x')|$, so $f^*(x)$ has the same strict sign as $f^*(x')$ and hence $y^*(x) = y^*(x')$. Together,

$$\hat{y}_\theta(x) = \hat{y}_\theta(x') = y^*(x') = y^*(x),$$

so $\{\hat{y}_\theta \neq y^*\} \subseteq \mathcal{X} \setminus U_\theta$. Since $\mathbf{1}\{\hat{y}_\theta(X) \neq Y\} \leq \mathbf{1}\{y^*(X) \neq Y\} + \mathbf{1}\{\hat{y}_\theta(X) \neq y^*(X)\}$, taking expectations under $P$ yields

$$\mathrm{err}_P(\hat{y}_\theta) \leq \mathrm{err}_P(y^*) + P_X(\hat{y}_\theta(X) \neq y^*(X)) \leq \mathrm{err}_P(y^*) + P_X(X \notin U_\theta).$$

If $\mathrm{supp}\, P_X \subseteq U_\theta$, the right-hand side vanishes. □

## B Miscellaneous

### B.1 Additional Information on Section 4.1

The networks trained on real and on synthetic data were trained with 100 seeds each. The conditional label distribution is shared by both marginals and is modelled as $p(1|(x_1, x_2)) = \min\{\max\{\sigma(x_2), \kappa\}, 1 - \kappa\}$ with $\kappa = 0.05$. We trained a MLP with two hidden layers of width 64 and ReLU activations, mapping to a single logit $z(x)$, which is bounded via composition with $f(x) = B\tanh(z(x)/B)$ with $B = \log((1 - \kappa)/\kappa)$. Assumptions ($L^\infty$) and ($\kappa$) are satisfied by construction. One model was trained on real (blue), one on synthetic samples (red) and both were evaluated against the same real test set.

### B.2 Details on Correlation Experiment

The tested models include variants of vanilla CNNs Lecun et al. (1998), ResNet He et al. (2016), DenseNet Huang et al. (2017), MobileNet Sandler et al. (2018), Inception Szegedy et al. (2015), squeeze-excite CNN Hu et al. (2018), dilated CNN Yu & Koltun (2016), ViT-small Dosovitskiy et al. (2021), MLP Rumelhart et al. (1986), MLP-Mixer Tolstikhin et al. (2021), random-wired networks Xie et al. (2019), spectral mixers Rao et al. (2023), Swin-inspired windowed attention (no window shifting) Liu et al. (2021) and involution networks Li et al. (2021), where hyper parameters like kernel size or number of layers were varied. All networks are trained from scratch on $3 \times 224 \times 224$ inputs. Depth, width, kernel size, patch size and attention head count (if applicable) were drawn randomly and each instance also draws one activation from {ReLU, GELU, SiLU, leaky ReLU, Mish, ELU}, applied uniformly throughout that network. The windowed-attention variant applies local window attention at a single resolution, without shifted windows, hierarchical patch merging, or relative position bias, so it is not a proper Swin Transformer. The spectral mixer takes only the global learnable frequency-domain filter from GFNet.

### B.3 Alternative Lipschitz bound

In order to obtain a bound that is in spirit more similar to Theorem 1 and Theorem 3 one can take the following angle.

**Theorem 12** (Convergence under a Lipschitz constraint)**.** *Assume* $(\kappa)$, $(L_f)$ *and* $(L_p)$, *and set* $L \coloneqq L_f + (B+F)L_p$. *Then for any* $\delta \in (0,1)$, *with probability at least* $1-\delta$ *over* $V_n \sim Q^{\otimes n}$,

$$\mathrm{err}_P\left(\hat{y}_{\hat{\theta}_n}\right) - \mathrm{err}_P(y^*) \lesssim \sqrt{\varepsilon_{\mathcal{F}} + r_n(\delta) + L\, W_1(P_X, Q_X)}\,,$$

*with* $r_n(\delta)$ *as in Theorem 3.*

*Proof.* By Lemma 11, we have $\mathrm{err}_P\left(\hat{y}_{\hat{\theta}_n}\right) - \mathrm{err}_P(y^*) \lesssim \left(\mathbb{E}_{P_X}\left[\mathrm{KL}_X\left(p \parallel p_{\hat{\theta}_n}\right)\right]\right)^{\frac{1}{2}}$. By Lemma 10, applied for fixed $V_n$, one has $\mathbb{E}_{P_X}\left[\mathrm{KL}_X\left(p \parallel p_{\hat{\theta}_n}\right)\right] \leq \mathbb{E}_{Q_X}\left[\mathrm{KL}_X\left(p \parallel p_{\hat{\theta}_n}\right)\right] + L\, W_1(P_X, Q_X)$, and by Theorem 1, with probability at least $1-\delta$ over $V_n \sim Q^{\otimes n}$, $\mathbb{E}_{Q_X}\left[\mathrm{KL}_X\left(p \parallel p_{\hat{\theta}_n}\right)\right] \lesssim \varepsilon_{\mathcal{F}} + r_n(\delta)$. Combining the three displays yields the stated inequality. □

Bounds of this form appear in domain adaptation, where the target risk of a Lipschitz hypothesis is controlled by the source risk plus a Wasserstein term (Redko et al., 2017; Shen et al., 2018, Lemma 1). These results bound an expected loss between two hypotheses, whereas we need the conditional Kullback–Leibler risk, and Redko et al. (2017) measure the Wasserstein cost in an RKHS rather than the Euclidean metric. It must be noted however that $\varepsilon_{\mathcal{F}}$ and $L$ are inversely related: As $L$ decreases $\varepsilon_{\mathcal{F}}$ increases. Therefore, the bound might be U-shaped. Also the Jensen and Pinsker step are very pessimistic so that in practice the bound in Theorem 7 should be preferred.

### B.4 Experimental validation of Theorem 7

For this experiment, we sampled $Q$ uniformly from the unit disk and $P$ uniformly from the concentric annulus $1 < \|x\| \leq 1.1$, which is disjoint from the disk. The dataset is depicted in Figure 6 (left). The shared concept map was modeled as $\eta(x) = \sigma(4x_2)$ clipped to $[0.05, 0.95]$. As neural network architecture we chose a 2-32-32-1 ReLU MLP. We enforce a Lipschitz bound $L$ on the network by projecting the singular values of each of the three weight matrices onto $\left[0, L^{1/3}\right]$ after every optimisation step, for $L \in \{0.5, 1, 2, 4, 8, 16, 64, \infty\}$. For each $L$, we train with Adam (learning rate $10^{-2}$, 2000 full-batch steps) on a weighted binary cross-entropy over 1000 synthetic points with labels drawn from $\eta$ and 1000 real points with adversarially flipped labels, weighting the real points by $\lambda \in \{0.01, 0.03, 0.1, 0.3, 1\}$. Among all checkpoints (every 40 steps) whose synthetic excess error is at $\varepsilon = 0.05$, we report the largest real excess error, i.e. the worst-case model that is still accurate on $Q$. We compare it to the bound given by the real mass not covered by the specified open balls around correctly classified synthetic points, estimated with 4000 real and 8000 synthetic samples. Excess errors are measured against Monte Carlo estimates of the Bayes error from 20,000 samples.

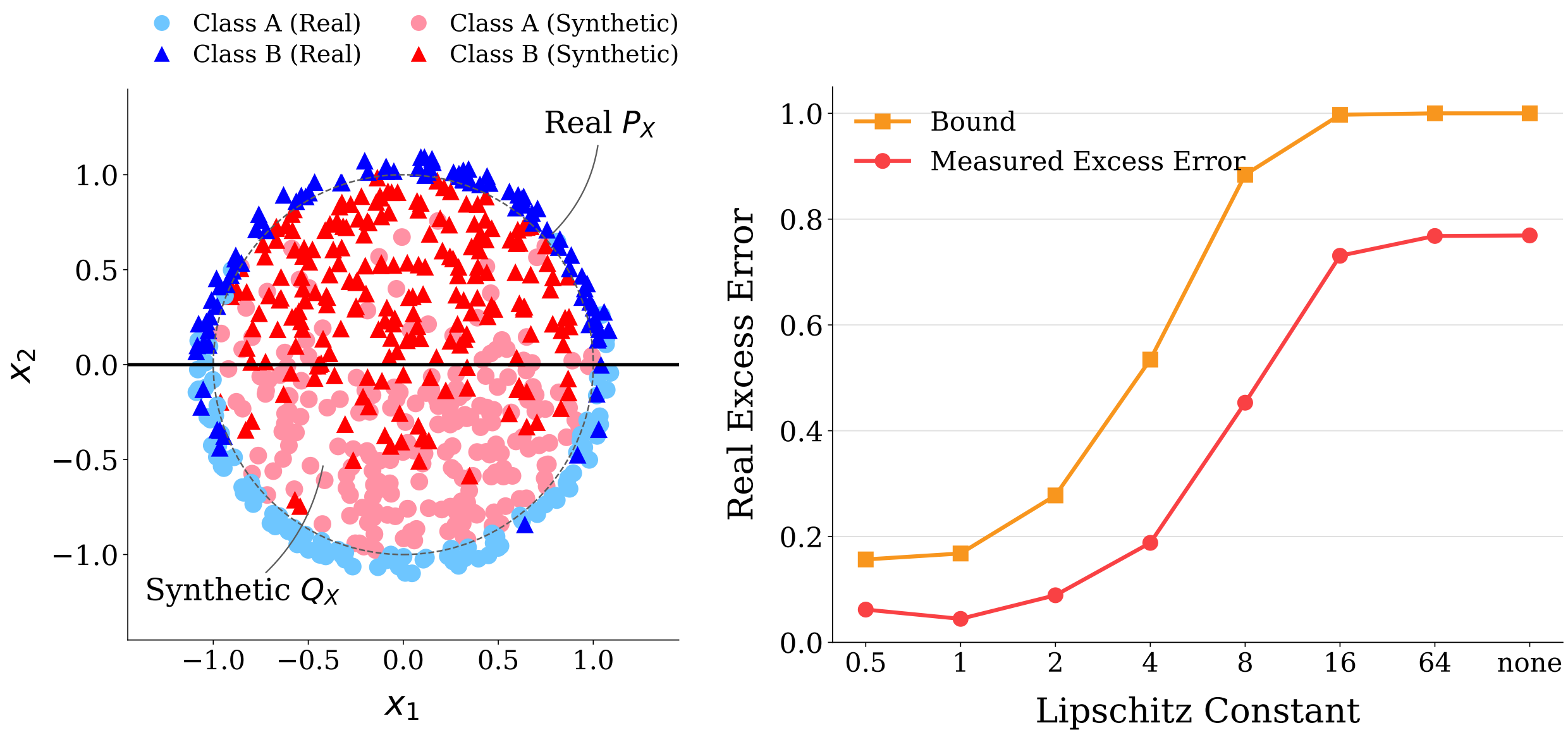

Figure 6: Dataset (left) and excess error against Lipschitz constant, cf. Appendix B.4